\documentclass[conference]{IEEEtran}

\usepackage[utf8]{inputenc}
\usepackage[T1]{fontenc}
\usepackage{amsmath,amssymb,amsfonts}
\usepackage{graphicx}
\usepackage{hyperref}
\usepackage[numbers,square,comma]{natbib}
\usepackage{url}
\usepackage{subcaption}
\usepackage{algorithm}
\usepackage{algpseudocode}
\usepackage{enumitem}

\algrenewcommand\algorithmicrequire{\textbf{Input:}}
\algrenewcommand\algorithmicensure{\textbf{Output:}}

\title{NFQ2.0: The CartPole Benchmark Revisited}
\author{\IEEEauthorblockN{Sascha Lange\thanks{Corresponding author: sascha@psiori.com}}
\IEEEauthorblockA{\\
Freiburg, Germany\\
Email: sascha@psiori.com}
\and
\IEEEauthorblockN{Roland Hafner and Martin Riedmiller}
\IEEEauthorblockA{Google DeepMind\\
London, United Kingdom\\
Email: rhafner@deepmind.com, riedmiller@deepmind.com}}

\begin{document}

\maketitle

\begin{abstract}

    This article revisits the 20-year-old neural fitted Q-iteration (NFQ) algorithm on its classical CartPole benchmark. NFQ was a pioneering approach towards modern Deep Reinforcement Learning (Deep RL) in applying multi-layer neural networks to reinforcement learning for real-world control problems. We explore the algorithm's conceptual simplicity and its transition from online to batch learning, which contributed to its stability. Despite its initial success, NFQ required extensive tuning and was not easily reproducible on real-world control problems. We propose a modernized variant NFQ2.0 and apply it to the CartPole task, concentrating on a real-world system build from standard industrial components, to investigate and improve the learning process's repeatability and robustness. Through ablation studies, we highlight key design decisions and hyperparameters that enhance performance and stability of NFQ2.0 over the original variant. Finally, we demonstrate how our findings can assist practitioners in reproducing and improving results and applying deep reinforcement learning more effectively in industrial contexts.

\end{abstract}

The neural fitted Q-iteration (NFQ) algorithm \cite{riedmiller2005neural} was the first algorithm successfully applying neural networks within the then-new ``fitted'' \cite{gordon1995stable,ormoneit2002kernel,ernst2005tree,lagoudakis2003least} paradigm of Reinforcement Learning (RL) to real-world control problems. Its success in these real-world applications and its increased stability-- when compared to asynchronous earlier variants of combining RL with neural networks--- can be considered the necessary groundwork for later algorithms such as DQN \cite{mnih2015human}, DDPG \cite{lillicrap2015continuous}, and MPO \cite{abdolmaleki2018maximum} and as one of the reasons for the increased interest in Deep Reinforcement Learning.

In hindsight, the most compelling aspect of NFQ might be its conceptual simplicity and, resulting from this, its ease of implementation: at its core is just the well-known vanilla Q-learning rule\footnote{Honoring standards in the targeted application domain, that is industrial automation, we use the cost variant with $c_{t+1} = c(s_{t}, a_{t}, s_{t+1})$, usually in the range of $[0, 1]$ where $0$ is best and $1$ worst.}
\begin{equation}
\begin{aligned}
Q(s_t,a_t) \leftarrow\  & (1-\alpha) \  Q(s_t,a_t)\ + \\
                      &\alpha\ [c_{t+1} + \gamma \min_{a} Q(s_{t+1},a)]
\end{aligned}
\end{equation}
computed along observed state transitions $(s_t, a_t, c_{t+1}, s_{t+1})$ and the representation of the Q-function's approximation in a multi-layer neural network. New was applying the ``fitted'' idea \cite{gordon1995stable} for storing the q-values in a neural network: up to then the Q-learning rule was used in an "online" or ``asynchronous'' fashion to update the neural network's weights immediately after seeing or ``replaying'' \cite{lin1992self} a state transition $(s_t, a_t, c_{t+1}, s_{t+1})$. In the ``fitted''-paradigm the neural network is updated ``synchronously'' for the whole state space to a new improved approximation of the Q-function by sweeping over all the observed transitions ``synchronously''. This change from an ``online'' to an ``offline'' or ``batch'' learning process is the main reason for NFQ's increased stability \cite{lange2012batch}.

With its conceptual simplicity, NFQ as presented in 2005, can be considered the vanilla version of modern Q-Learning in Deep Reinforcement Learning.

Despite its success in practice on simulated as well as on real-world control problems and its ability to learn good controllers, it also introduced a lot of new hyperparameters and questions that were not all well understood at the time. Thus, observed results were not easily reproducible
and the learning process was still not very robust and repeatable, in the sense of consistently improving the learned policy from episode to episode and producing the same good policy after the same number of episodes from run to run. Rather, NFQ applied to real-world control problems still required a lot of tuning, careful design of state and action spaces and the manual selection of good controllers from the sequence of learned policies of varying quality.

Subsequent work in Deep Reinforcement Learning build on the fitted idea and applied it to continuous action spaces \cite{hafner2011reinforcement}, to higher-dimensional state spaces like visual input \cite{lange2010deep,mnih2013playing} and also to higher-dimensional action spaces \cite{abdolmaleki2018maximum}. With the complexity of problems and tasks, the complexity of the algorithms also increased.

Nowadays, algorithms from the ``growing batch'' family \cite{lange2012batch}  such as  DQN, DDPG, and MPO, are implemented in the most popular RL libraries and frameworks and work well out of the box on popular (simulated) benchmark tasks, also involving visual input, learning repeatably and robustly good control policies.

Comparing the first experiences with the ``early'' algorithms on technical systems like the CartPole, inverted pendulum, and acrobot and the more recent success of DQN, DDPG, and MPO and such on more impressive tasks, the question is, why are these newer algorithms more successful on even larger dimensional problems? Is it because the ``core algorithms'' (the temporal difference updates, the usage and execution of the Bellman operator, the approximation of the value function) are so much better or do other advances and learnings outside the core algorithms contribute as well or even more? Specifically, beyond the quality of resulting policies, why has the learning process with the old algorithms been so much less robust, fragile and not easily reproducible? What makes today's algorithms so much more robust and easier to apply?

To get some insights into these questions, we will revisit the CartPole task,

concentrating on the harder (compared to the simulated benchmark tasks) physical real-world system, built from standard industrial components,
with a modernized version of the original algorithm NFQ and compare the results to its original variant.
We will investigate the learning process itself, its repeatability, robustness and its possible pitfalls more thoroughly than has been done before, in order to better understand, what made learning so hard in the early days as well as why practitioners new to the field still have so much trouble recreating results on their own real systems and transferring algorithms to different problems. We will then perform several ablation studies to understand what design decisions and hyperparameter choices cause the observed improvements over the original NFQ. Building on these results and insights, we will finally demonstrate how to modify and further improve the learned policies to create even better solutions to the real-world control problem.

We assume that the insights and findings form the analysis of the learning process, the ablation studies, and the step-by-step improvements of the learned policy will help practitioners not only to reproduce results on benchmark tasks such as the real-world CartPole, but also to help them transfer algorithms to different problems and industrial systems and to understand our own analysis, reasoning and choices made while overcoming the problems and pitfalls of the system setup as well as the NFQ algorithm as published in the original paper.

It turns out that the original, patent-free NFQ algorithm, modernized with nowadays knowledge to NFQ2.0, is still a very good and competitive algorithm and can be significantly improved regarding learning results, speeds, and robustness by simple modifications and hyperparameter choices. This aligns with the common wisdom that knowing and mastering your tool, in industrial contexts, is often more important than a small \textit{potential} advantage in performance promised by a different, new, and seldom less complex algorithm.

NFQ2.0 as well as the configurations and parameterizations used in this paper are available as open source. NFQ2.0 is easy to use and it is useful for a broad range of real-world control problems.

We'll close the report with a discussion of the findings and its implications for the application of Deep Reinforcement Learning in industrial settings.
A better understanding of these questions regarding the hyperparameters and contributions of the different techniques and ideas should guide practitioners to apply Deep Reinforcement Learning more widely and successfully in this domain.

\section{Methods}

\subsection{Learning Algorithm}

For our evaluations, we used an unmodified 3rd party open source python implementation of NFQ that is publicly available\footnote{psipy, downloadable from \url{https://github.com/psiori/psipy-public}} under a permissive BSD-3 license. It implements the pseudo code of algorithm \ref{algo:nfq_main}, which is the same as originally published in \cite{riedmiller2005neural}, but offers much more flexibility in terms of the choice of hyperparameters and the setup of the learning task and also implements many of the tricks and tips introduced by Riedmiller in the book chapter \cite{riedmiller2012tricks} and Hafner and Riedmiller in \cite{hafner2011reinforcement}.
\begin{algorithm}[h]
\caption{NFQ Main Loop}
\begin{algorithmic}[1]
\Require {Set of transition samples $D$}
\Ensure {Q-value function $Q_N$}
\State $k \gets 0$
\State {\Call{init\_MLP}{} $\rightarrow Q_0$}
\Repeat
    \State {\Call{generate\_pattern\_set}{} $\rightarrow P \newline
       \hspace*{2.5em}   = \{(\text{input}_t, \text{q-target}_t), t = 1,\ldots,|D|\}$ where:}
    \State $\text{input}_t = (s_t, a_t)$
    \State $\text{q-target}_t = c_{t+1} + \gamma \min_{a} Q_k(s_{t+1}, a)$
    \State \Call{Rprop\_training}{$P} \rightarrow Q_{k+1}$
    \State $k \gets k + 1$
\Until{$k = N$}
\end{algorithmic}
\label{algo:nfq_main}
\end{algorithm}

The loop of the algorithm alternates between the duality of the Q-function representations as either a neural network $Q_k$ or in the form of sampled, Bellman-updated training patterns $P$. First, new, improved approximations of the individual q-values $Q_{k+1}(s_t, a_t)$ (``q-target'') of all observed state-action pairs $(s_t, a_t)$ are computed. This computation is done ``along'' all
observed transitions $(s_t, a_t, c_{t+1}, s_{t+1})$ by adding the observed reward $c_{t+1}$ and the expected q-value $Q_k(s_{t+1}, a)$ of the ending state $s_{t+1}$ according to the current approximation $Q_k$. In literature, this step is commonly called the "Bellman update", ``Temporal Difference step'', short ``TD'' step, or ``Dynamic Programming'' or short ``DP'' step. Then, a neural network is trained on the resulting fixed set $P$ of training samples $((s_t, a_t), q{\text{-target}_t})$ using standard supervised learning, here in the batch mode variant using RProp as the optimizer. This step produces an improved (neural) approximation $Q_{k+1}$ of the q-function and is commonly referred to as ``fitting'' or the ``fitted'' step.

We speak of a duality of representations here, as both the neural network as well as the set $P$ of state-q-target tuples carries all the information
presently available to the iterative learning algorithm and both of them are enough to derive a policy, individually. After calculating the new q-targets, the current neural approximation is discarded and a newly initialized neural network is fit on the data.\footnote{Yes, nowadays we usually continue training an existing neural network on the dataset $P$, but in the old days, Riedmiller and others used to train a newly initialized neural network from scratch in each iteration, throwing away the previous neural approximation after calculating the new q-targets.} After fitting the neural approximator, the old q-targets are discarded and new q-targets are calculated. Furthermore, there are algorithms like the kernel-based algorithm of Ormoneit and Sen \cite{ormoneit2002kernel} that don't use an explicit representation of the approximated q-function at all. Instead, they do all calculations directly in the sample space and also derive a policy directly from the q-targets $P$ at query time following a non-parametric, lazy learning approach.

\begin{table*}[bt]
    \centering
    \begin{tabular}{|l|c|c|c|}
    \hline
    \textbf{Feature} & \textbf{NFQ (2005)} & \textbf{NFQ2.0} & \textbf{DQN} \\
    \hline
    Network Topology & small, 20x20  & larger, 256x256x100 & Deep CNN \\
    \hline
    Action in... & input neuron & input neuron & output neurons \\
    \hline
    Activation & tanh + sigmoid & ReLu + tanh + sigmoid & ReLu + linear \\
    \hline
    Weight initialization & uniform -0.5 to 0.5 & Glorot uniform & ? \\
    \hline
    Training Strategy & Batch updates & Mini-batch & Mini-batch \\
    \hline
    Optimizer & RProp & Adam & RMSProp \\
    \hline
    Loss Function & MSE & MSE & MSE \\
    \hline
    Normalization & unit-variance & unit-variance & ? \\
    \hline
    Gamma & 1.0 & 0.98 & 0.99 \\
    \hline
    Approximator strategy & re-train from scratch & keep & keep \\
    \hline
    Terminal goal states & Yes & No & No(?)  \\
    \hline
    Epochs per Bellman update & 300 / till convergence & 8 & ? \\
    \hline
    Bellman updates per episode & 1 & 4 & ? \\
    \hline
    Exploration ($\epsilon$-greedy) & constant 0.1 & linear schedule $0.8-0.05$ & linear schedule $1.0-0.1$ \\
    \hline
    Tricks & at least one of hint-to-goal, x++, or min-subtraction & none & clipped rewards \\
    \hline
    \hline
    \multicolumn{4}{|l|}{\textbf{Task / CartPole specific choices}} \\
    \hline
    Cost function & sparse & sparse / shaped & -  \\
    \hline
    Stacking (real CartPole) & 1.5 (previous action added \cite{riedmiller2012tricks}) & 6 & - \\
    \hline
    Pole position & angle & (cosine, sine) & - \\
    \hline
    \end{tabular}
    \caption{Comparison of key algorithmic differences between NFQ, NFQ2.0, and DQN. For details on the choices, see the text.}
    \label{tab:algo_comparison}
\end{table*}
The main loop of NFQ works on a fixed set of transitions, that was sampled from the environment. In reality, NFQ and derivatives were and are almost always used in a growing-batch variant \cite{lange2012batch,riedmiller2012tricks}, where the set of transitions is not fixed, but grows over time and where the current controller is used to sample new, more interesting transitions for the next iteration. Whereas pure random sampling can lead to good controllers in simple tasks like e.g. the original pendulum balancing task, this incremental collection of experience is necessary to solve more complex tasks, such as the swing-up task, because random sampling of transitions would be unlikely to see good trajectories through the state space or would never reach the desired goal state at all.

A whole set of important parametric choices is hidden in the innocent-looking RProp\_training function: besides the "obvious" supervised learning hyperparameters as choice of input and output representation, network topology, activation functions, learning rate, and weight decay, there is also the more subtle but rather impactful choice of how many epochs to train the network on the batch of samples. Extremes from initializing a fresh network and training till convergence to continuing with the same network as used in the last iteration and only updating the weights a single time can and have been chosen.

Whereas the core algorithm of NFQ in its simplicity is still sound and valid, the parametric choices that could be made at the time of its invention-- from today's perspective-- were rather constrained by computational feasibility and the lack of better methods and practices in neural network training.

Thus, the main differences between now (NFQ2.0) and then (NFQ) are in the choices ``outside'' the high-level pseudo code and in the chosen hyperparameters (such as gamma, network topology, activation function, weight initialization, training strategy, optimizer, etc.) and in the setup and modeling of the learning task (such as cost function, input and output representation, stacking, etc.). Table \ref{tab:algo_comparison} provides an overview of these key differences between the original NFQ algorithm, our modernized variant NFQ2.0, and DQN. As the original paper's author has not provided a public implementation, we have collected the subtle design choices and somewhat hidden hyperparameters from the available literature \cite{riedmiller2005neural, riedmiller2005pole,riedmiller2012tricks,lange2012batch,hafner2011reinforcement} and from personal communication. The modifications we list for NFQ2.0 reflect the advances and learnings in supervised deep learning practices and reinforcement learning techniques that have emerged over the span of more than a decade.

We made the following choices and modifications for NFQ2.0:
\begin{itemize}
    \item we use larger networks with wider and deeper hidden layers as the original NFQ suggested. For the CartPole and similar tasks, Riedmiller suggested 20x20 hidden neurons as the default, where we now suggest 256x256x100 neurons.\footnote{We suggest to use networks of a magnitude wider than the size that was proposed with the original NFQ for the CartPole and a larger suite of benchmark problems. As the optimal network size is usually application dependent, we cannot provide a specific topology recommendation that is optimal for all applications. However, for industrial (non-visual) data and up to a few dozen discrete actions we suggest three layers with 256x256x100 as a default network size and a good starting point to optimize from, if topology search is desired and feasible. Whereas Hafner and Riedmiller made a point in keeping the network size small, even smaller than 20x20 in some applications \cite{hafner2011reinforcement}, we found that even larger networks are often easier to train (less iterations in supervised learning) and might offer some performance improvements. But, in our experience, measurable differences are so negligible on many real industrial systems, that the additional computational effort is often not worth it. Significantly smaller networks, on the other side, can be harder to train and may lead to significantly worse policies, as is demonstrated in the ablation studies.}
    \item we use ReLu activation functions for their speed in the first hidden layers, tanh in the feature layer (last hidden layer, which is always fully connected) and the sigmoid in the output layer for the q-values.
    \item we initialize the network weights using Glorot's uniform initialization scheme that scales the weights to fan-in and fan-out automatically, as suggested by \cite{glorot2010understanding}.
    \item we use Adam as optimizer with the default parameters.
    \item we use mini-batches with a default size of 2048 samples.
    \item we alternate between collecting a full episode of data and training the network on the collected data for either a given number of episodes or until no further improvement in the average costs per step is observed.
    \item we train a single network from start to finish of an experimental run, not re-initializing its weights after collecting more data or re-calculating new q-targets.
    \item after every Bellman step, we update the network weights for 8 epochs towards the new q-targets.
    \item after every time the batch was grown with an explorative episode, we do 4 Bellman steps to speed-up "back-propagation" of newly obtained information along the trajectories.
    \item we normalize the input features to have zero mean and unit variance automatically every 10 episodes for half the training run, then keep the scaling fixed for the rest of the training. See section \ref{sec:ablations} for a discussion.
    \item we also scale the action values to the range $[-1, 1]$.
    \item we make extensive use of stacking (concatenating the n previous actions and states as additional input features) to provide the network with more information about the system dynamics and the past to counter latencies on real systems.
    \item we use $\gamma = 0.98$ as discount factor, as the default, as it is the more robust choice also for infinite-horizon tasks than using $\gamma = 1.0$, as originally advocated by Riedmiller in \cite{riedmiller2005neural,riedmiller2012tricks}. $\gamma=1.0$ assumes either a finite-horizon task, certain bounds on the expected future costs, or the careful application of other tricks to keep the approximated Q-function from diverging.
    \item we go with shaped cost functions as the default, as they help guide the exploration and speed-up learning. We present techniques for switching to different cost functions during the training procedure on the fly, including the usually preferred time-optimal cost function, in section \ref{sec:evaluation} and section \ref{sec:improvements}.
    \item we use a linear $\epsilon$-schedule as the default, which starts with a high value of $\epsilon=0.8$ and decays linearly to a minimum of $\epsilon=0.05$ within a quarter of the total training episodes. This high randomness is definitely not needed when using shaped costs, but we chose the slow change from rather high to low $\epsilon$ because it has proven to be very robust and reliable in many different setups even when other settings were far from optimal in our preliminary studies. For a further discussion of the impact of the $\epsilon$-schedule, see section \ref{sec:ablations}.
    \item we use none of the tricks from \cite{riedmiller2012tricks}, not because we question their usefulness, but because it turned out that we don't need any of them when using $\gamma < 1$ and either a shaped cost function or an $\epsilon$-schedule starting with very high randomness. When using $\gamma = 1.0$ at least one, if not several, of the listed tricks would be needed, alone, in order to avoid (early) divergence of the Q-function.
\end{itemize}
Mini-batch size, and the relation between Bellman steps, network epochs and explorative episodes have been determined empirically in preliminary experiments using different simulation setups. It is not the optimal solution for the CartPole or any other system, but a good compromise between speed of learning, computational efficiency, and robustness of the learning procedure in many different setups. Whereas Bellman updates spread new information backwards along trajectories from one location back to previous location in the state space (only one step at each iteration), fitting moves the neural network's estimations ``in location'' (at the starting state of a transition) closer to the ``local'' q-target which was improved by the Bellman update. More Bellman updates can spread new information faster, more fitting steps can move estimates faster to significantly changed q-targets. But, there is a trade-off here between number of needed system interactions and the number of Bellman updates and network training updates, which are potentially wasted, when no ``informative'' data was collected at the start of the learning process or in a particular explorative episode. Thus, the interplay between the number of Bellman updates and the number of fitting steps per Bellman update is delicate and can easily increase the computational burden at little benefit when chosen too high. Riedmiller used 1 Bellman update per episode and training for 300 epochs (due to using a fresh network after each episode of data collection) \cite{riedmiller2012tricks}, DQN was used with either 1 or 2 training epochs per Bellman step in the Atari papers, which demonstrates the drastic differences in parameters that still lead to good learning results.

We would adapt the mini-batch size if the length of the episodes and goal-leading trajectories would be significantly different from what we have and use on the CartPole (less than 100 steps to reach the goal, episodes of 200 to 400 steps).

We tried to make NFQ2.0 the best variant of NFQ / DQN, having the architectural choices and parameters of NFQ on the one (old) end and the choices of DQN at the other (more modern) end, testing both and choosing the better one in the sense of a) quality of found policies and b) robustness of learning procedure. If unsure (contribution negligible and / or below significance thresholds), we went for the simpler choice, aiming for the most simple, yet (near) optimal solution.

\subsection{System}

\begin{figure}[hbt]
    \centering
    \includegraphics[width=\columnwidth]{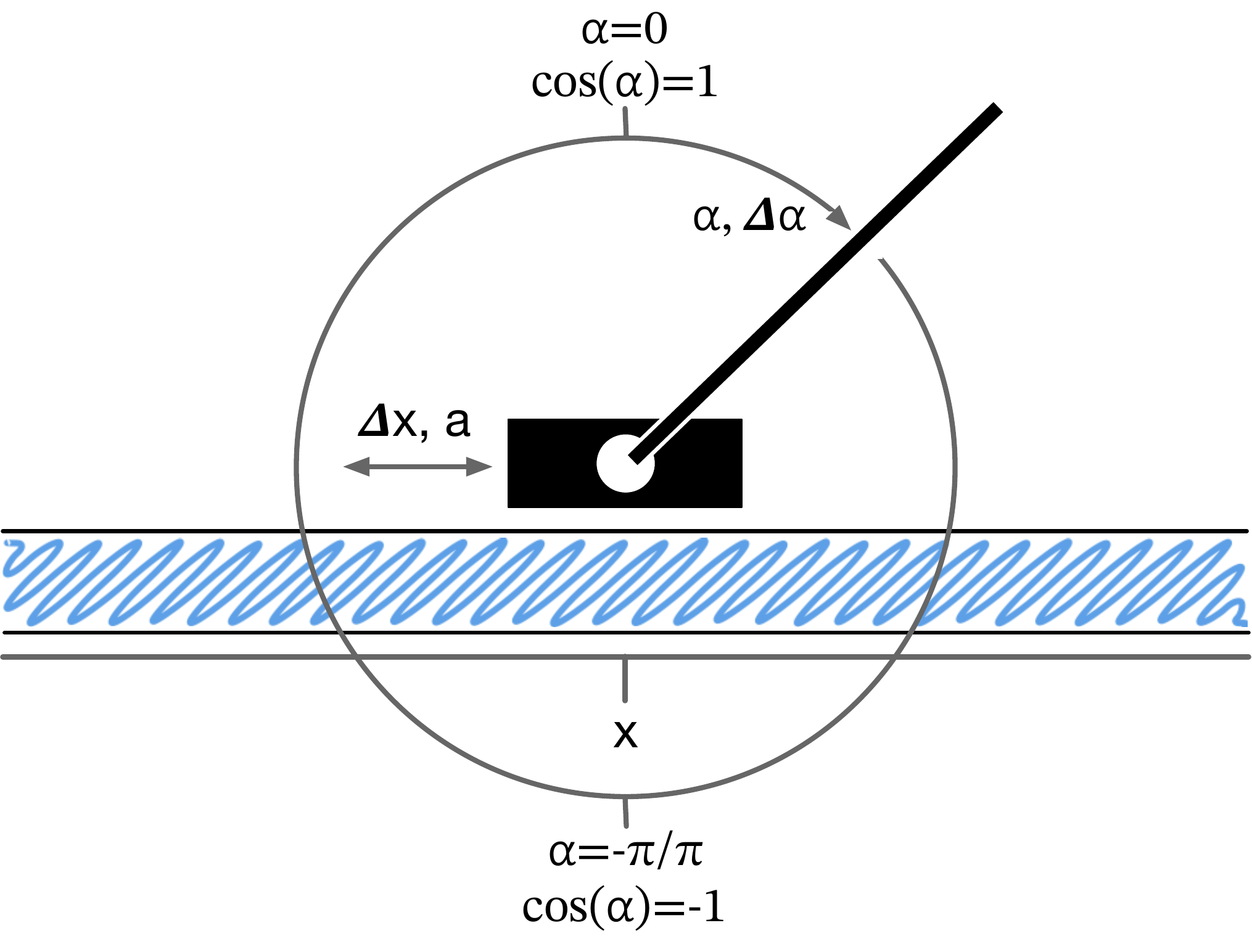}
    \caption{Schematics of the CartPole system used for the evaluation of NFQ2.0. The movement of the cart can be influenced by an action $a$ that is either a force applied to the cart (simulation) or a target speed setpoint for the cart (real system). A pole is attached to the cart and can swing freely around the pivot point. The task is to swing up the pole from a downwards hanging position and balance it indefinitely while keeping the cart in the center area of the track. The cart's position $x$, its velocity $\Delta x$, the pole angle $\alpha$ and its angular velocity $\Delta \alpha$ are observed, on the real system with the help of two encoders, one mounted on the pivot point and one attached to the band pulling the cart. The pole angle $\alpha$ in this setup has a non-linearity "jumping" from $-\pi$ to $+\pi$ moving through the downward position. The angle is often replaced by the pair $(sin(\alpha), cos(\alpha))$ in the state representation, in order to avoid this discontinuity and achieve a continuous representation.}
    \label{fig:cartpole_system_schematics}
\end{figure}
For the evaluation of NFQ2.0 as well as for the ablation studies, we used the swing-up and balance task (see fig. \ref{fig:cartpole_system_schematics}) on both, a simulated CartPole system as well as on a real, physical system. Whereas the simulated system was used for preliminary tests and the design of some of the experiments, the real system was used for all evaluations and the ablation studies presented in this report. We chose this focus on the real system, as the original papers mainly focused on the simulated benchmarks, not going into too much detail about the harder real system. Furthermore, the real system is the one that poses several challenges, that are not present in the simulated system, such as delays, noise, and restrictions on the available data, etc., which make the application of Deep RL methods more interesting and challenging and often hinder a wider adoption of the methods in industrial control.

The simulation comes together with the NFQ implementation in the psipy package.
It is a fork from the OpenAI Gym / Farama Gymnasium CartPole environment that was extracted some time ago from Farama's version 0.29.1 (August 2023) and that was modified to also support the swing-up task by not ending automatically when letting the pole drop below +/- 24 degrees as is hard-coded in the Gymnasium version. Default parameters for pendulum length, gravity, etc. remained unchanged.
\begin{figure*}[tb]
    \centering
    \includegraphics[width=\textwidth]{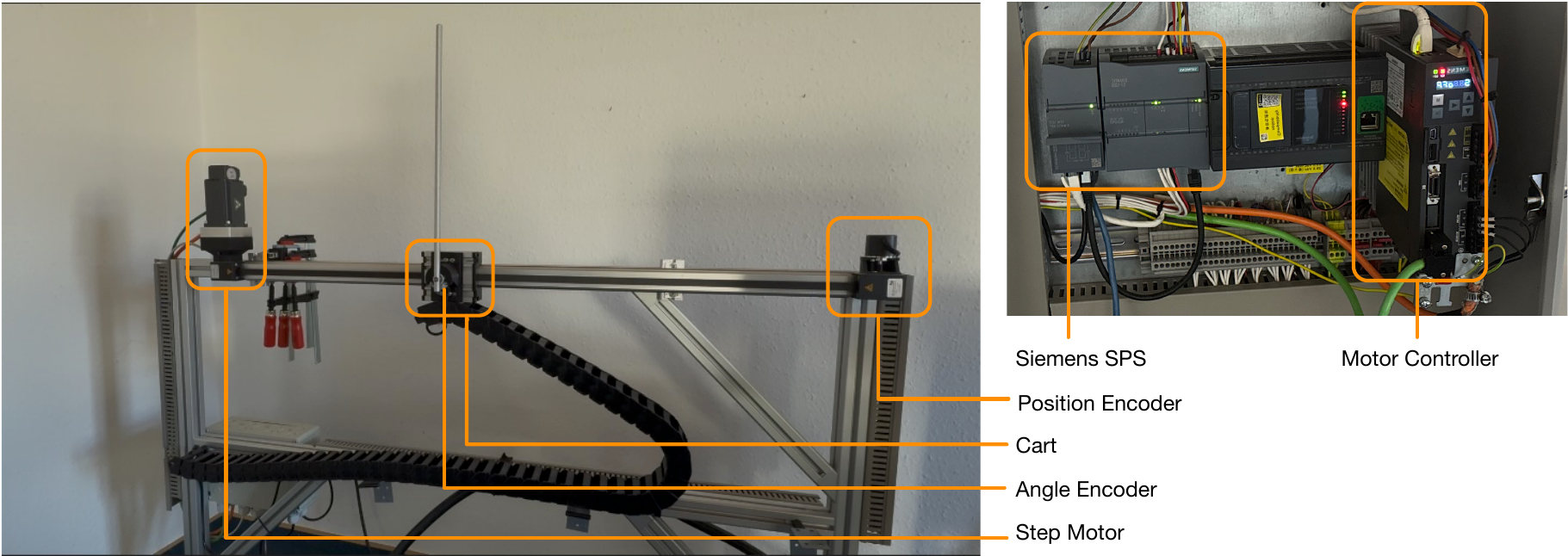}
    \caption{The real CartPole system for evaluation of NFQ2.0. The system is a  commercial product provided by a 3rd party vendor. It is build from non-modified, off-the-shelf industrial automation standard components that are widely used in real plants doing mass production. Therefore, it provides the necessary robustness and durability for doing extensive testing and evaluation of the learned controllers without human oversight. Furthermore, it resembles systems that can be found in real plants, for the goods (standardization, durability) and bads (limits and inflexibility due to black-box and standardized components, communication bloat, latency build-up) and is therefore a very good starting point for the development of learning-based methods and controllers to be used in real plants. The CartPole system consists of a linear actuator driving a cart on a linear rail with the help of a band and a pulley system. A servo motor is use to drive the band and the cart. One encoder is mounted on the pivot point and one is attached to the band pulling the cart. There is one physical switch mounted in front of the motor that is used for the calibration of a zero-position of the cart. The encoders are directly connected to a Programmable Logic Controller (PLC), a Siemens S1200, via a High Frequency (HF) input module. The actuator is driven by a specialized Motor controller, a Siemens V90, that is connected to the PLC via ProfiNet, an Ethernet-based industrial protocol and fieldbus. An industrial computer, also provided by the vendor, in the IPC-form factor (not depicted) is used to run the learning algorithm and the neural control algorithm in real-time. We have chosen the Intel-based option, without any specialized components (no TPU / GPU), because we didn't expect any significant benefits from running the very small models on a GPU, but drawbacks considering CPU and bus performance from the NVIDIA Jetson-based alternative offering. The system runs an Ubuntu LTS. It comes equipped with a specialized interface card offered by Hilscher and a small C++ component providing an easy-to-use ZMQ API for user programs to communicate with the CartPole System via ProfiNet.
    The PLC has been programmed using Siemens's proprietary programming language, STEP 7 via their software "Tiaportal". We use a vendor-provided minimal program that realizes basic logic for implementing endstops (left side with the switch, right side virtual, depending on encoder reading) to protect the system from damage. }
    \label{fig:cartpole_system_real}
\end{figure*}
The real system, as depicted in fig. \ref{fig:cartpole_system_real}, is a  commercial product provided by a 3rd party vendor.

The original system used in \cite{riedmiller2005neural} was specifically optimized for providing low-latency control from the Linux machine running the learning algorithms. Running at 20 Hz everything was designed to let the agent collect the effects of a chosen action in cycle $t$ already with the cycle $t+1$, meaning there where less than 50ms to process an incoming state, chose an action, send the action to the CartPole system, apply it to the motors, let it be executed, collect the changes from the sensors and communicate everything back to the Linux machine to start the next cycle. All this was designed to come as close as possible to the ideal case where the action starts to effect the system immediately at cycle start and the effect is measured over the whole cycle and measured and communicated with zero-latency in the exact moment where the new cycle starts. This is especially important for NFQ, as the  theoretical foundation of the algorithm relies on the system state representation to be Markov, meaning that the current state contains all information about the past and future that is needed to make an optimal decision.

The vendor of the CartPole system and the authors share a common experience: this level of careful design and control performance is never met in an existing industrial system, and seldom can be achieved in the wild. Most industrial systems are not designed to respect the Markov property and, in reality, often rely on a chain of standardized components plugged together, with more or less care for latency, timing and jitter. While Programmable Logic Controllers (PLCs, industry's variant of ``hardened'' micro controllers) usually advertise high control frequencies of 100 Hz or more, you seldom find these in real implementations, and, if found, these are somewhat invalidated as full loops of measurement-control-effect-measurement often take several cycles of the chained component's internal control loops, already in the communication, because of each components own internal control loop and (unknown) timing.

Thus, the design choice for the CartPole system as well as for our experimental setup has been to mimic the less-than-ideal reality of industrial systems as much as possible, using non-modified off-the-shelf industrial components for the hardware of the CartPole system. The vendor provides an Ubuntu LTS--based control computer that is directly connected to the ProfiNet fieldbus of the PLC on the CartPole system. On the control computer runs a small vendor-provided C++ component named ``busy'' that receives the state information from the field bus as they become available, places the control commands on the bus at known intervals and establishes a synchronous, time-stamped control loop and a user-friendly ZMQ-communication channel for user space applications, such as ours, run on the same computer.

\begin{figure}[hbt]
    \centering
    \includegraphics[width=\columnwidth]{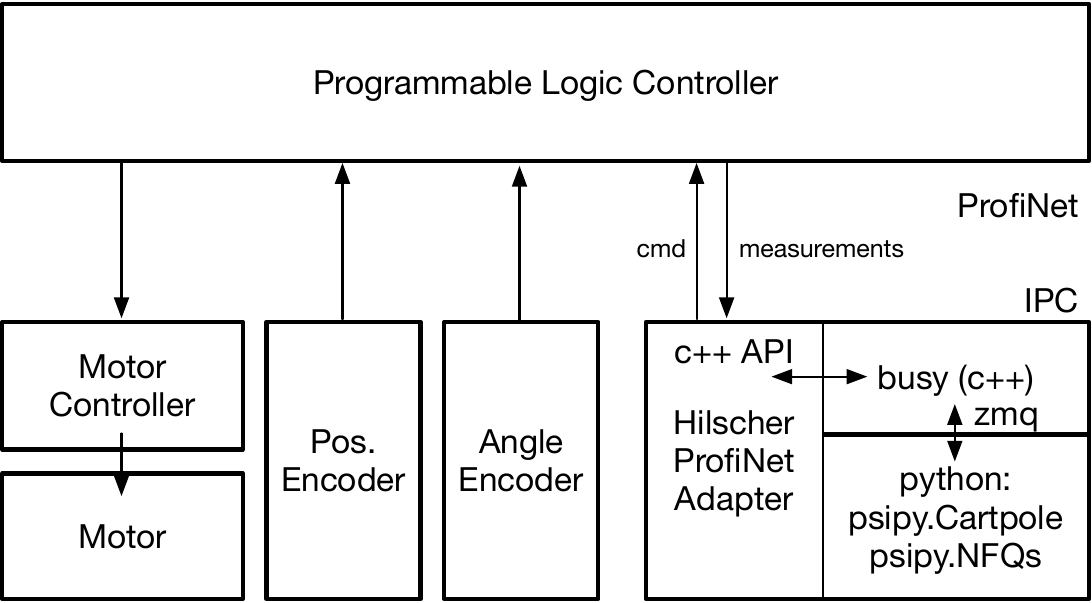}
    \caption{The full line of communication between the control computer and the CartPole system. Each box represents a physical component of the system that communicates with the Programmable Logic Controller (PLC) as the central component via ProfiNet, an Ethernet-based fieldbus standard. Within the control computer (IPC), several hardware (Hilscher ProfiNet Adapter) and software components (busy, python user program) participate in the communication. The PLC, the motor controller, busy and the psipy-based user program each establish an internal control loop, reading and writing data. Our user program syncs its internal control loop to busy's control loop, which is responsible for establishing a jitter-free control loop, where the sensor information is provided to the user program at the start and the next command is send at the end of the control loop, preferring a known, constant and jitter-free interval over the lowest possible latency (when sending the command immediately after it becomes available). The other components can't assume to establish any synchronization among each other and run internally at different frequencies of 50 Hz or more. When running busy and NFQ at 20 Hz, the overall latency is about 2 cycles, around 100ms, measured by the time needed for a given set-point change / flank, having any effect to the received encoder measurements. Internal unknown states of the motor controller, such as e.g. a build-up integral component, have been observed to have even later effects after a given user command (set point change).}
    \label{fig:cartpole_system_real_communication}
\end{figure}
The psipy package provides a ready-to-use CartPole environment for this physical system. It handles the communication with the systems C/C++ component "busy" running on an IPC and provides exactly the same interface as the simulated one, so that the same learning algorithm can be used for both, simulated and real system. The full line of communication is depicted in fig. \ref{fig:cartpole_system_real_communication} and---from the viewpoint of a RL-researcher--- sadly, reflects the state of the art in industrial control systems quite adequately.

\subsection{Metrics}

During the evaluations, we will usually report and plot the learning curve on basis of the average costs per steps observed during an episode. We use the average instead of the total costs, in order to account for different episode lengths, either because of the experimental setup or because of reaching the termination criteria by reaching the endstops of the physical CartPole or hitting the track-bounds in the simulation.

Since the cost function is part of the system design and setup, and therefore an integral part of the Reinforcement Learning approach, it may vary between learning experiments and does not necessarily provide meaningful comparisons to other methods from control theory. Therefore, for evaluating the quality and performance of the learned policies, we utilize additional, more objective metrics that are independent of the system design.

Specifically, we calculate the following stability and deviation metrics on individual episodes as introduced in \cite{hafner2011reinforcement}:
\begin{itemize}
    \item $N$: as the number of time-steps after the controlled process variable enters the tolerance range around the set-point (and does not leave it again).
    \item $e^{\infty}$: as the mean absolute off-set the controlled process variable has from the set-point after a certain number of time-steps $N_{max}$
    \item $e_T$:  as the mean absolute off-set that the controlled process variable has from a predefined reference trajectory
\end{itemize}

We add another metric $n$ to measure whether or not and how fast a controller can at least swing up the pole, if not stabilize, and balance it, by defining
\begin{itemize}
    \item $n$: as the number of time-steps after the controlled process variable enters the tolerance range around the set-point for the very first time.
\end{itemize}
Differing from $N$, the controlled variable may leave the tolerance range again afterwards.

We use the zero angle of the pole as the reference trajectory during the balancing phase and calculate $n$, $N$, $e^{\infty}$ and $e_T$ on the pole angle alone, ignoring the cart position. We define the tolerance area as $+/-$ 10 degrees around the zero angle. With this setup, $N$ basically describes how fast a controller is able to swing up the pole and stabilize it not leaving the tolerance area ever again. $e^{\infty}$ and $e_T$ are calculated after $N_{max} = 200$ steps, as most good policies manage to swing up and balance by this time. For policies where N turns out to be larger than $N_{max}$, we calculate the stability metrics starting from step $N+20$, to give the policy 20 steps time to stabilize after entering the tolerance area.

The metric $N_{cart}$ is almost always smaller than $N_{pole}$. Therefore, we will not report $N_{cart}$ and cart position metrics in the following.

For controlling and evaluating the supervised learning process of the neural network, in our case, looking at the MSE is unfortunately not very helpful, as it becomes rather small quickly and then moves a lot because of the targets shifting with every Bellman update. It only helped us to identify situations, where nothing was learned at all, e.g. due to a very badly chosen learning rate, mini-batch size, or too small network topology. What helps a little more, is looking at the minimum, average, and maximum of either the q-targets or the q-values, which are simply the network's output. We usually plot min, avg, and max of the latter for each epoch while training. For the CartPole task, these values should be spread widely from each other, the maximum somewhat close to the terminal costs, the minimum close to zero, as soon as the agent ``understands'' there is a near-zero stable solution, after stabilizing the pole. The average should stay somewhere between those two values. In bad setups, neither of these values spreads at all, in good setups, the maximum will spread quickly, while minimum and average stay closer together, until the first near-zero transition costs are collected.

\section{Evaluation}
\label{sec:evaluation}

We evaluated the performance of NFQ2.0 on the swing-up and balance task on the real system.\footnote{All code used for running the exepriments and producing the graphs and results is available at \url{https://github.com/salange/nfq2-experiments}.} We'll discuss the setup of the learning and evaluation process, the learning curves, results and several aspects regarding robustness and repeatability of the learning process in the following.

\subsection{Setup of the learning and evaluation process}
\label{sec:setup_of_learning_and_evaluation_process}

For the CartPole task, we use the cart's position $x$, its velocity $\Delta{x}$, the pole's angle $\alpha$ encoded as cosine and sine  as $(cos(\alpha), sin(\alpha))$ and its angular velocity $\Delta{\alpha}$ in the state representation
\begin{equation*}
    s=(x, \Delta{x}, cos(\alpha), sin(\alpha), \Delta{\alpha})
\end{equation*}

We designed the cost function\footnote{We use the cost framework everywhere, due to customs in control theory. It is equivalent to the reward framework-- where in the cost framework we minimize costs, we'd maximize rewards.} for both, the simulation and the real system, as follows:
\begin{equation*}
    c(s_{t+1}) = \left\{
    \begin{array}{ll}
    1.0 & \text{if } x \in x_{\text{- -}} \\
    0.05 & \text{if } x \in x_{\text{-}} \\
    0.01 \cdot (1-\frac{\cos(\alpha)+1}{2}) & \text{if } |x - \text{c}| < \theta_{\text{cart}} \\
    0.01 & \text{otherwise}
    \end{array}
    \right.
\end{equation*}
where the center of the track $c$ is defined as track-length/2 and $\theta_{\text{cart}}=0.15 \cdot \text{track-width}$ is a tolerance margin on the target cart position in the middle of the track. $x_{- -}$ and $x_{-}$ both denote regions outside the desired working area, where $x_{--}$ is the area of the hard endstops, giving maximum costs of 1 and ends the trajectory, which leads the transition into $s_{t+1} \in x_{--}$ to be treated as a terminal transition, meaning that in the the TD-update step in algorithm \ref{algo:nfq_main} we use $q_{\text{target}}(s_{t+1}, a) = c(s_{t+1}) = 1.0$ instead of adding the expected future costs as well. $x_{-}$ is a slightly wider area among the endstops, which we introduced as a ``soft-stop'' area, to take away strain from the hardware system. Entering $x_{-}$ leads to higher than usual costs, but the episode is not terminated. Regarding an industrial application, $x_{--}$ would be considered as a hard constraint the system imposes, either physically or by safety measures in the underlying control system, thus, where entering would be considered a process failure, with potential emergency stop or other more dire consequences. $x_{-}$ in this case would be part of the learning system design and would be chosen to discourage the controller from entering areas which are either running risk of failure, e.g. close to the endstops, or known to be better avoided and not helpful at all. See \cite{riedmiller2012tricks} for details on the concept behind x-work / x-minus and x-plus.

Within the working area the agent always receives constant standard step costs of 0.01\footnote{The relation between terminal costs (1.0) and step costs (0.01) is important, as it must be cheaper to stay on the track and to continue than to end the episode intentionally early, to avoid any further future costs after collecting the terminal costs once. Our choice is safe, because $\sum_n^{\infty} 0.98^n = 50$ and, thus, a rather uninformed, but safe policy would expect not more than $50 \cdot 0.01 = 0.5 < 1.0$ future costs.} except for a area in the center of the track. In this area, the costs are ``shaped'', meaning they establish a cost gradient towards the goal state, where lower costs are collected the closer the pole is to the upright position. For this, we use the $cos(\alpha)$ of the pole angle $\alpha$ as a shaping function, scaled in such a way that it is exactly the normal step costs of 0.01 for the pole angle $\alpha=\pi$ (downward position) and 0 for the pole angle $\alpha=0$ (upright position).

The actions $a$, in the default setup, are the applied cart forces from a set $a \in (-10, 0, 10)$ Newton in simulation and velocity setpoints from a set $a \in (-300, 0, 300)$ communicated to the motor driver in the real system.

In the simulation, we experience no latency and the actions are chosen and applied ``instantly''. Thus, no stacking is needed.

On the real system, we experience latencies of about 100ms and more, thus, more than the span of two full control cycles. This clearly violates the Markov-condition and renders the system non-Markovian. Thus, Q-learning is not applicable to the state information $s$ as is. Therefore, we treat the system as a Markov Process of higher order, where the state $s$ is extended to include the last $n$ actions $a_t, a_{t-1}, \ldots, a_{t-n+1}$ and states $s_{t-1}, s_{t-2}, \ldots, s_{t-n+1}$ as additional input channels. In the default setup, we use a history-length or ``lookback'' (term used in the implementation) of $n=6$ for the real system, including the last 6 states and actions, counting the current ones. From an preliminary analysis of the system response to simple on-off ``flanks'' (abrupt, large setpoint changes send to the motor controller) we found that $n=6$ is generously accounting for all observed latencies and delayed effects. Generally speaking, its better to err on the long side than on the short side, as including more information than necessary does not harm in theory, whereas to err on the short side would lead to a non-Markovian system model, which would, at best, let some effects non-explainable and, thus, non-learnable by the controller, or, at worse, lead to a complete failure to learn anything but an ``average'' or random behavior. A deeper analysis of the actual history length's influence on the learning process can be found in section \ref{sec:ablations}.

With this setup, we ran NFQ2.0 5 times in independent training runs for 500 episodes with 400 steps (or until termination in an endstop) each. Our reasoning was to give the controller enough time to swing up during a single episode, even with initial imperfect strategies or by random chance. In hindsight, 200 steps per episode likely would have been more efficient, as the balancing phase is already reached within 70 steps or less by good policies. After each training episode, we ran a single greedy ($\epsilon=0$) evaluation episode to calculate the average cost per step of the present policy, also for 400 steps (or until termination in an endstop). Plotting and reporting average costs was chosen over plotting total costs in order to account for episodes of different length specifically in the beginning of the learning phase (if unaccounted for the number of steps executed in an episode, total costs of a bad, full episode in a plot could be visually close to a very short episode where the agent crashed immediately in the endstop). The transitions observed in these evaluation episodes were discarded and not used for training.

The very first episode is started with the cart near the middle of the track, the pole hanging down. All subsequent episodes are started where the previous episode ended, or, if an endstop was hit, at the initial state near the middle of the track.

\subsection{Learning curves}
\label{sec:learning_curves}

\begin{figure}[thb]
    \centering
    \includegraphics[width=\columnwidth]{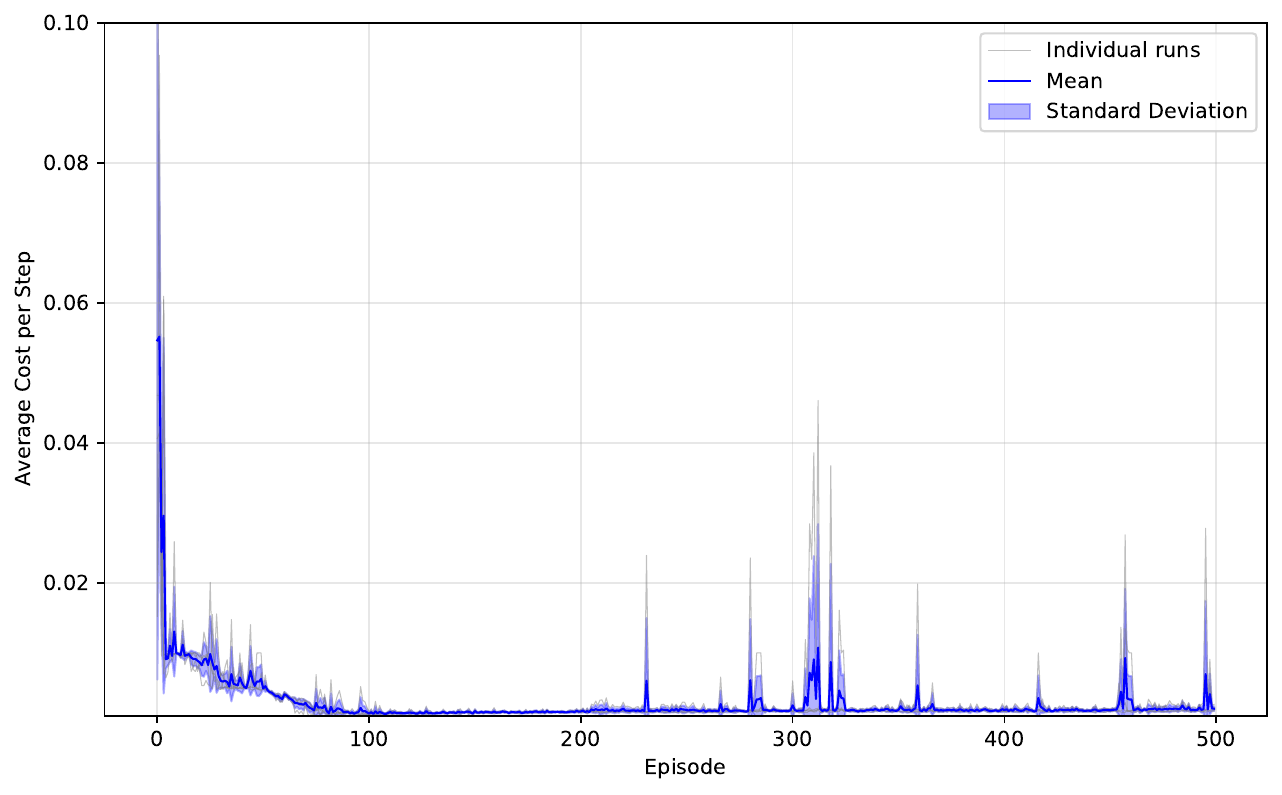}
    \caption{Performance of NFQ2.0 on the CartPole task. Plotted is the average cost per step in a single distinct evaluation episode (greedy evaluation, data not used in training) after n episodes of training. All runs learn a good policy within 120 episodes that manages to swing up and stabilize the pole reliably. Learning curves show consistent improvement towards the best learned policies with only minor stochastic regressions and surprisingly little variance among the individual runs. Thick line: average over 5 independent training runs. Thin lines: individual runs. Blue area: standard deviation.}
    \label{fig:learning_curve_real_cartpole}
\end{figure}
Figure \ref{fig:learning_curve_real_cartpole} shows the learning performance of NFQ2.0 in terms of average costs per step collected during evaluation episodes. The learning process is quite smooth, with the system steadily reducing costs over time with only minor "stochastic" regressions. Proper policies which reliably reach the goal state are typically found within the first 50 to 70 episodes, while good policies that maintain the pole upright for the entire duration after reaching the goal state once are usually found within 70-120 episodes. In all 5 runs, the best policy, defined as the one with the lowest average cost in an evaluation run, was discovered already within the first 93 episodes in average, with no further improvements afterwards regarding the average costs.

Notably, the individual runs demonstrate very similar learning progression and results, evidenced by the low variance in the learning curves, especially in the later phase from episode 80 to about 200. This demonstrated stability and the repeatability of the learning process is noticeably better than during the initial experiments with NFQ in the 2000s.\footnote{Since the learning curves haven't been published in the original papers, we have to rely on personal communication with the original authors and the results of the later ablation studies, especially the ones with smaller network topologies and those which  re-initialized the network weights every episode, to make this point.}

Within each of the training runs, we could identify several phases and milestones the policies went through during their improvement. We list them here to help replicating our results and to understand the learning process better:
\begin{description}
    \item[learning to avoid endstops]
    the first few episodes usually end in the endstops. Typically, during a full training run, we hit them between half a dozen and a dozen times, more often, when we don't use the soft-stop area.
    \item[jittering safely]
    the endstops are avoided safely, the agent typically alternates between left and right, learning to stay more in the middle of the track, avoiding the hard and soft stop areas, collecting first costs $< 0.01$. The maximum q-values already spread from the min and avg, indicating the agent ``understood'' that not all actions lead to a terminal state, eventually.
    \item[increasing energy] the agent starts increasing the energy on the pole, swinging it higher and higher. This typically starts within the first 30 episodes.
    \item[swing-over] the pole is swung over the top doing a full circle. This is an important milestone that was found within the first 40 episodes of each of our 5 test runs, but, that in some suboptimal setups (seeing the ablation studies) is found significantly later after hundred(s) of episodes or not at all. After achieving this once, minimum q-targets and average q-targets immediately start to spread, whereas before this, they stayed rather close to each other.
    \item[windmill] the pole is swung over the top and rotated with constantly increasing, quite impressive speeds testing the physical limits of the system. The windmill is a phase that almost all our training runs went through. In the early days, we feared this phase and tried to avoid it, because the impression was, many learning processes got stuck in it, as the local optimum leads rather to higher rotational speeds, than to the zero speed needed in a successful balancing phase. Nevertheless, in all experiments conducted for this report, all controllers that reached the windmill phase also learned to pass it, eventually, although it took some setups mimicking the ones in the original paper 200 episodes or more to do so. Waiting longer instead of stopping a run early probably would have been enough also in the past.
    \item[catching phase] the pole is caught after the swing up and held for a short moment, perhaps a second, upright. If it falls down the side where it came from, this is a good sign for a quick improvement in the next episodes.
    \item[balancing] the pole is successfully balanced the whole episode, only dropped by random actions due to epsilon-greedy exploration. The deviation from the upright position is improved steadily and then gets stuck at a certain level not improving any further. The swing-up phase becomes more aggressive and efficient from more random starting positions. In this phase we observe the minimal q-targets to approach values close to zero, one or two orders of magnitude below the standard step costs of 0.01. Collecting more samples from the balancing phase, the average q-targets settle around the standard step costs of 0.01. This value is highly dependent on the length of the episodes and the gamma parameter chosen, of course.
\end{description}

120 episodes amount to about 48000 transitions ($120$ episodes times $400$ steps minus the "missing steps" from the--- usually less than a dozen--- early terminated episodes), which amounts to about 2400 seconds or 40 minutes of interaction with the system. The complete training time with evaluations and learning updates takes less than 2 hours up to this point.

Although we did neither optimize the episode length nor the exploration strategy to achieve best results within the fewest interactions possible, even our average results of needing 93.2 episodes to find the best policy match or beat the results reported in literature for model-free approaches in a similar task setup (e.g. \cite{riedmiller2005neural}, \cite{riedmiller2012tricks} reporting 392 episodes of 200 steps in average).

The whole dataset of 500 episodes contains up to 200k transitions, needs two times of about 10000 seconds or 166 minutes of interaction with the system for learning and evaluation. The total wall time for such a run is about 16 hours, as the time needed for data handling and calculating the learning updates constantly increases with the number of transitions.

To save wall time in further evaluations and the ablation studies, we will evaluate the learning performance only on the first 200 episodes, where the best policies can be found reliably, from here on, and do long runs and multiple repetitions of the same setup only where necessary to make a point.

\subsection{Quality of the learned policies}
\label{sec:quality_of_learned_policies}

\begin{figure*}[t]
    \includegraphics[width=\textwidth]{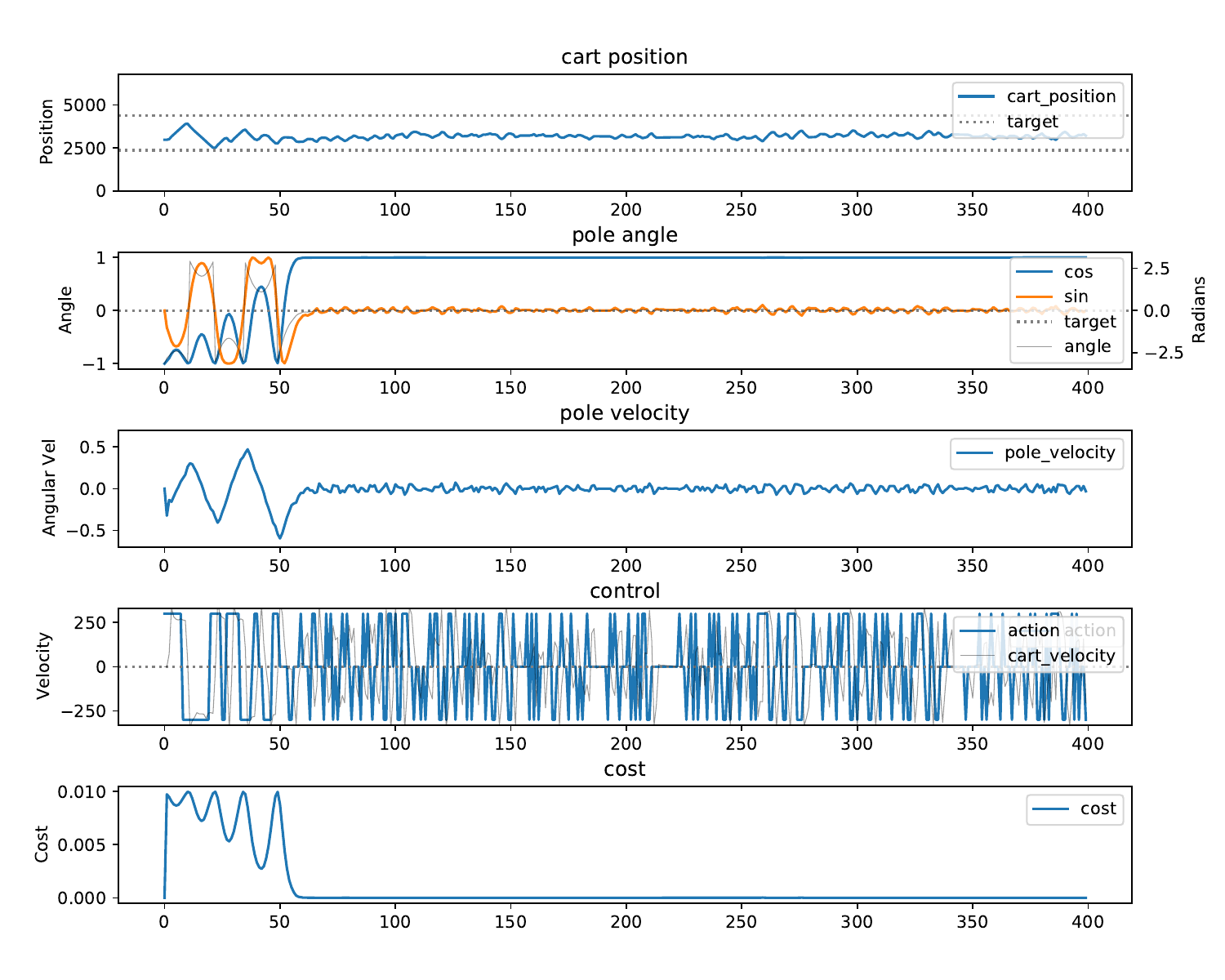}
    \caption{Evaluation of the policy with the lowest average cost in run 5 after 74 episodes. The pole is in the upright position and the cart is at the center of the track after about 60 steps.}
    \label{fig:best_policy_evaluation_episode_1}
\end{figure*}
Table \ref{tab:performance_metrics} shows the performance metrics of the policies with the lowest average costs in their evaluation run. These policies have reached the tolerance area and remained there after about 60 to 70 steps. In the balancing phase, the maximum deviation $e_T$ from the upright position was 6.0 degrees, the averaged deviation $e_{\infty}$ was between 1.25 and 1.69 degrees. Exemplary, figure \ref{fig:best_policy_evaluation_episode_1} shows the evaluation episode 74 of the training run 5, which was the one with the lowest average cost in all training runs.
\begin{table}[htb]
    \centering
    \begin{tabular}{l|c|c|c|c|c}
        \hline
        \textbf{Run} & \textbf{1} & \textbf{2} & \textbf{3} & \textbf{4} & \textbf{5} \\
        \hline
        \hline
        Episode & 108 & 107 & 80 & 97 & \textbf{74}  \\
        \hline
        $n$ & 69 & 70 & 58 & 72 & \textbf{59}  \\
        \hline
        $N$ & 69 & 70 & 61 & 72 & \textbf{59}  \\
        \hline
        $e^{\infty}$ & 1.60 & \textbf{1.25} & 1.67 & 1.69 & 1.60  \\
        \hline
        $e_T$ & 5.4 & \textbf{4.8} & 6.0 & 5.4 & 6.0  \\
        \hline
        Avg Cost & 0.0012 & 0.0012 & \textbf{0.0010} & 0.0011 & \textbf{0.0010}  \\
        \hline
    \end{tabular}
    \caption{Performance metrics across five independent training runs of NFQ2.0, showing the metrics of the best policy found in each run, where "best" is defined as the one with the lowest average cost per step in their single evaluation run. Whereas the the policies found in run 5 and 3 swung up the pole the fastest, and, because the costs during the swing up phase dominate the average costs, have the overall lowest average costs, the policy found in run 2 is the best one in terms of stability, as it has the lowest average deviation of $e_{\infty}=1.25$ in the balancing phase. Values for stability and deviation are given in degrees.}
    \label{tab:performance_metrics}
\end{table}

\subsection{Variance in the evaluation performance and selection of the best policies}
\label{sec:variance_in_evaluation_performance}

Are these policies really the best ones learned? Selecting the best policy based solely on the average cost per step measured in a single evaluation run is likely suboptimal, since the evaluation process is highly stochastic due to varying initial conditions and inherent randomness in the physical system's dynamics.
\begin{figure}[hbt]
    \centering
    \includegraphics[width=\columnwidth]{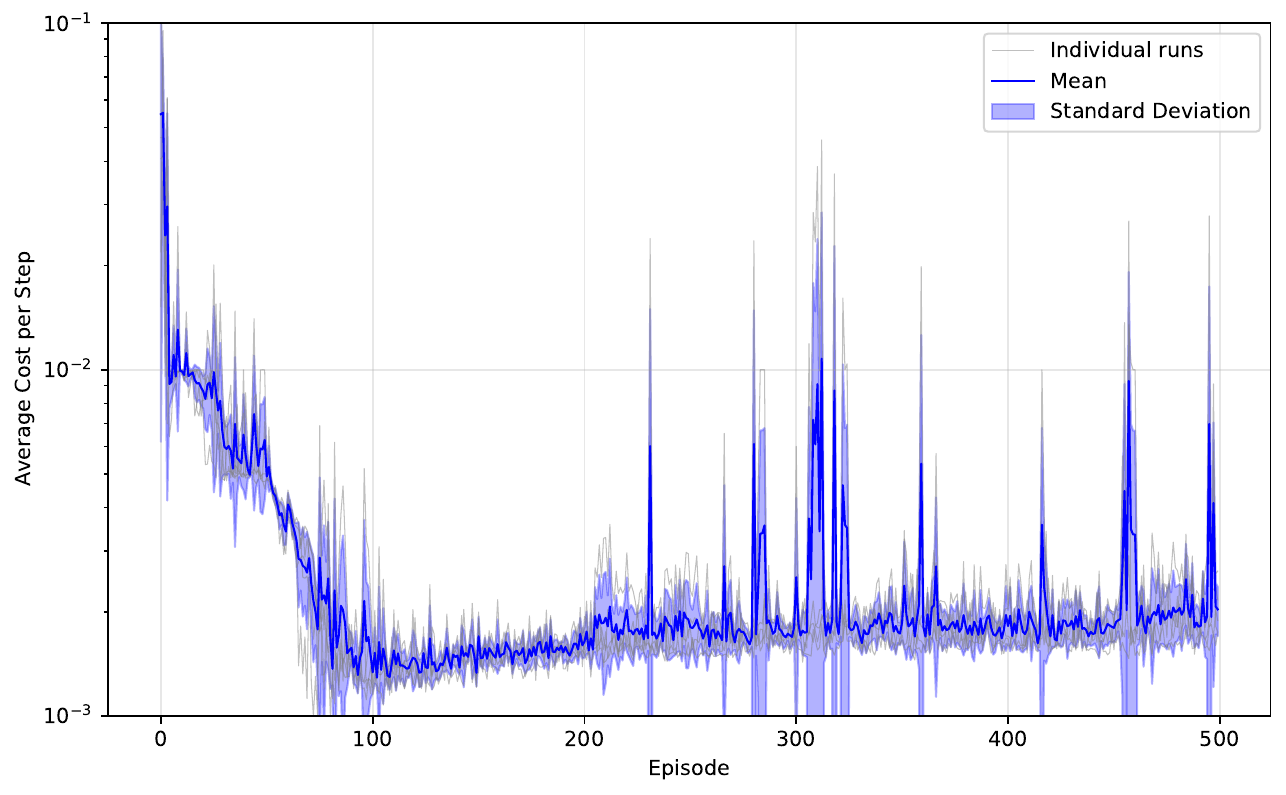}
    \caption{Performance of NFQ2.0 on the CartPole task. Plotted is the average cost per step in a single distinct evaluation episode (greedy evaluation, data not used in training) after  episodes of training. Thick line: average over 5 independent training runs. Thin lines: individual runs. Blue area: standard deviation. Left: Linear scale. Right: Zoom into the variance of the later learning process using a logarithmic scale on y-axis.}
    \label{fig:performance_log}
\end{figure}

Furthermore, the setup of the learning and evaluation process causes another bias in the average step costs measured in the evaluation, in this case preferring earlier policies: Figure \ref{fig:performance_log} depicts the same learning curve as before, but zoomed-in on a log scale. It can be seen, that the performance degrades slightly, but clearly visible from about 120 episodes to 200 episodes. This happened in all 5 runs. Further investigation shows that this is due to starting the evaluation where the learning process left the cart, just after finishing the update of the approximated q-function. As this updating process takes longer and longer with the number of episodes, the free-swinging pole loses more and more energy that might have been left from the previous learning trial, thus, making the initial state for the next evaluation run likely harder over time. Around 200 episodes the training update takes so long that the pole has practically no energy left whatever the situation the previous trial ended in.

Practical considerations---such as system availability, reliability, or the impossibility of controlling initial states---often prevent sufficiently extended and distinguishing evaluation after each individual training episode. This was also the case in our setup, where evaluation after each episode was limited. In these cases, it is good practice to select the best policy from an extended evaluation which is run on the best candidate policies found by a feasible, smaller evaluation during the training phase. If epsilon controlling the random exploration is sufficiently small during the training phase, relying on the average cost during training itself, is often also a valid strategy to select candidate policies for further evaluation.

In our case, epsilon is not small enough to avoid evaluation runs during training completely. Even an epsilon as small as $\epsilon=0.05$--- that is, one random action about every 20th step--- causes the pole to fall down regularly in the 20s training runs, even for policies that in evaluation runs show perfectly good stability with no blunders over several minutes. Thus, on our system, we can only distinguish stable candidate policies from unstable policies with a distinct greedy evaluation run.

Hence, we further analyzed the performance of other candidate policies of run 5, which we selected on basis of low average costs but also good values for $N$ and $e_{\infty}$ in their individual evaluation runs. In practice, selection can either be done by visually inspecting the plots of the evaluation runs or, computationally, by ranking the policies based on their performance metrics $N$ and $e_{\infty}$ in their single evaluation runs.

For the selected candidate policies we ran 5 additional, longer evaluation runs of up to 4800 steps (4 minutes) from several random starting positions with low to no energy left in the system, to increase the confidence in the evaluation results with a reasonable additional (wall clock) time investment.

\begin{table*}[htb]
    \centering
    \begin{tabular}{l|c|c|c|c|c||c|c}
        \hline
        \textbf{Episode} & \textbf{74} & \textbf{105} & \textbf{134} & \textbf{160} & \textbf{206} & \textbf{134-off} & \textbf{500-off}\\
        \hline
        \hline
        $n$ & $\mathbf{70.40 \pm 3.72}$ & $70.40 \pm 9.07$ & $71.40 \pm 10.31$ & $76.20 \pm 18.86$ & $71.40 \pm 9.85$ & $75.60 \pm 7.14$ & $63.80 \pm 10.11$ \\
        \hline
        $N$ & $75.60 \pm 11.32$ & $844.80 \pm 1446.18$ & $\mathbf{71.40 \pm 10.31}$ & $91.00 \pm 27.20$ & $77.80 \pm 22.34$ & $75.60 \pm 7.14$ & $68.80 \pm 14.63$ \\
        \hline
        $e^{\infty}$ & $1.68 \pm 0.07$ & $1.48 \pm 0.05$ & $1.24 \pm 0.06$ & $\mathbf{1.23 \pm 0.05}$ & $1.27 \pm 0.10$ & $1.22 \pm 0.07$ & $1.19 \pm 0.08$ \\
        \hline
        $e_T$ & $6.72 \pm 0.70$ & $ 5.76 \pm 0.29 $ & $5.76 \pm 0.29$ & $5.64 \pm 0.29$ & $\mathbf{5.04 \pm 0.48}$ & $4.92 \pm 0.45$ & $4.32 \pm 0.96$ \\
        \hline
    \end{tabular}
    \caption{Extended evaluation metrics for different candidate policies from run 5, showing mean and standard deviation across multiple evaluation runs. Also looking fine in its original single evaluation run during the training phase, policy 105 let the pole fall down after balancing it for a while in several of the additional evaluation runs, which resulted in the high value and variance of $N$. The best overall policy learned during the growing-batch training phase according to this evaluation is the one after 134 episodes, which features both, the best stabilizing behavior $N$ and--- at least when considering the variance bounds--- also the shared best stability $e_{\infty}$, meaning it swings up and stabilizes the pole the fastest and shows the best stability during the balancing phase. Policy 160 shows a similar stability (average $e_{\infty}$ is non-significantly better by 0.01) but a less stable swing up behavior. The last two columns (134-off) and (500-off) show the performance of two policies learned purely offline, from the dataset of the first 134 episodes as well as on the full dataset of 500 episodes. The offline-policy trained on the full dataset demonstrated the overall best stability and a non-significant better swing-up behavior.}
    \label{tab:extended_evaluation_metrics}
\end{table*}
Table \ref{tab:extended_evaluation_metrics} shows the performance metrics of the original candidate (74) and 4 later candidate policies after learning for 105, 134, 160 and 206 episodes, in these extended evaluations, together with the observed variances.

It becomes clear, that the best policy is not necessarily the one that performed best in the single evaluation run during the training phase. Whereas the variance of the repeated measurements in the balancing phase are small, if not negligible, the variance in the swing-up phase is significantly larger and, obviously, also more heavily dependent on the initial starting conditions.

Specifically, policy 74, also in these evaluations, is still the one that reaches ("touches") the tolerance area the fastest but its advantage over the other policies is not as large as the single evaluation run let it seem ($59$ steps in the single evaluation run compared to the more reliable measurements of an average of $70.4$ steps in the extended evaluations). Considering the more important metric $N$, the number of steps needed to stabilize the pole in the tolerance area, it's not the fastest swing-up policy anymore. On the other hand, the stability metric $e_{\infty}$ was only slightly underestimated in the single evaluation run, showing only slightly worse performance in the extended evaluations ($1.60$ in the single evaluation run compared to $1.68$ in the extended evaluations). Generally, the variance measured in the estimates of $e_{\infty}$ between individual repetitions of the same policy from different starting states is quite small, with values of $0.07$ to $0.1$ degrees.

Given the extended evaluation, with increased confidence, but not absolute certainty, we consider the policy after 134 episodes the best one for further evaluation and all comparisons, as it features both, the best $e_{\infty}$ (shared within the variance bounds with policy 160) and $N$, meaning it swings up and stabilizes the pole the fastest and shows the best stability at the same time. Policy 206 comes close, as the worse value in $N$ is caused by a single bad trial and might be a random coincidence.

\subsection{Eventual unlearning after about 200 episodes}
\label{sec:eventual_unlearning}

In all 5 runs we observed that eventually the policies performance degraded noticeably. Whereas the vast majority of policies from about 100 to 200 episodes manages to stabilize the pole in the tolerance area and never leave it in the evaluation runs, also not leaving the tolerance area of the cart position after stabilizing the pole, the majority of policies from 200 to the end of the training phase either leave the tolerance area of the pole angle regularly, let it fall down or collect high costs due to leaving the center position of the track. The resulting learning curve on a log scale is somehow similar to the effect of overfitting in supervised learning. Neither more data nor more Bellman interactions do help at least not within 500 episodes. As we don't want to overfit on the problem at hand, for example by optimizing a schedule of learning parameters of the neural network, we leave further investigation of this effect for future work and chose to stop the training phase at 200 episodes in the following experiments, at least if we don't see any improvements in the evaluation metrics approaching this point.

The learning curve \ref{fig:performance_log} also features a few spikes of high costs after 200 episodes. Looking at the evaluation runs, these spikes are caused by the policies failing to swing up the pole at all or by hitting the endstop, which also causes high costs. Most of these episodes have in common, that they were starting from a state already very close to the endstop area. Additional evaluations of these policies from several random starting states did not show any particularly bad performance, but a performance close to that of previous and subsequent policies. This indicates that these spikes were rather not caused by particularly bad policies, but by bad luck in the starting positions in the evaluation run, unfortunately, hinting at the learned policies  not being perfectly reliable close to the endstop area. If starting from here would be a requirement of a hypothetical application, exploring with the chosen reward shaping might be counter-productive and should be either changed, to not lead the policies away from the endstop area too early but let it explore the full working area more extensively or chose a different strategy to initialize the starting position of the training runs, enforcing regular starts and a more reliable behavior close to the endstops.

\subsection{Offline learning versus exploring while learning}
\label{sec:offline_learning_vs_exploring}
We use the growing batch approach because with a pure batch-approach and collecting data completely random, the goal state would be reached only very rarely, if at all.

Now, after finding the best policy after 134 episodes, we have a dataset of useful episodes, which include the goal state as well as good trajectories through the working area reaching it.

Can we learn from this dataset a competitive policy from scratch, without letting it explore ever again and how long does this take, in the sense of TD updates?  We know, that the data was sufficient to learn a very good policy.

To address this question, we ran NFQ2.0 on the collected dataset of 134 episode from scratch, not letting it explore ever again, but evaluating its learning progress after the same number of TD updates (4) as during the original training run to allow us to analyze the learning process in detail. Since we start with a large amount of data and, from the original interactive learning runs, we know that at some point NFQ runs into overfitting-like behavior, we chose a learning rate one magnitude smaller than the one used in the original run (0.0001 instead of Adam's default 0.001) to reduce the risk of overfitting and at least extend the stable phase of the learning process.

\begin{figure}[thb]
    \centering
    \includegraphics[width=\columnwidth]{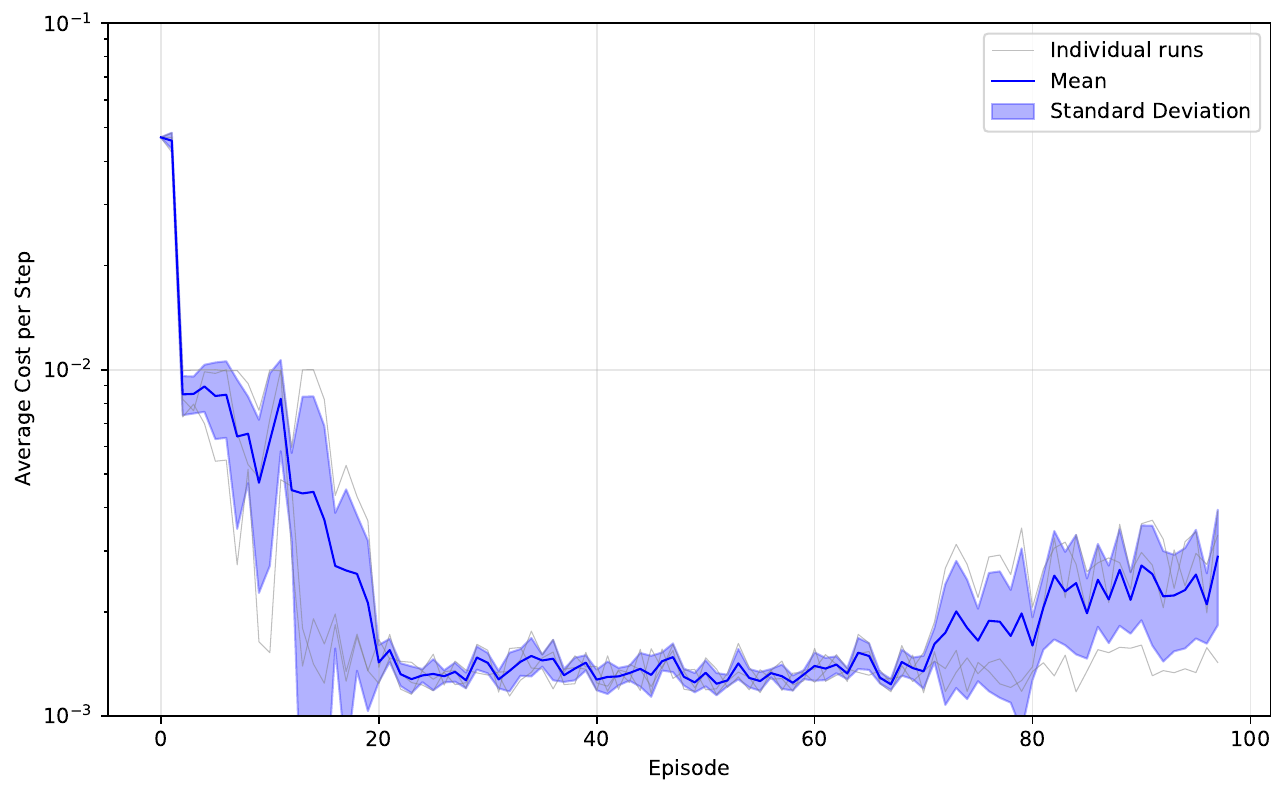}
    \caption{Performance of NFQ2.0 in pure offline training on the dataset of the first 134 episodes of the original training run 5. We ran 1 evaluation episode after every 4 offline TD updates of the neural network. The learning curves of three independent repetitions from different random initialized neural networks are plotted (light gray), as well as the average over all runs (blue) and the standard deviation (blue area). The learning rate was chosen one magnitude smaller than in the original run (0.0001 instead of 0.001) to reduce the risk of overfitting and at least extend the stable phase of the learning process. Evaluation was done after every 4 TD update, which is identical to the number of TD updates used in the original training run after every explorative episode.}
    \label{fig:learning_curve_real_cartpole_offline_log}
\end{figure}
Figure \ref{fig:learning_curve_real_cartpole_offline_log} shows the learning curves of three independent repetitions from different random initialized neural networks. We evaluated the neural network once after every four TD updates. All three runs reach a similar good performance after about 64 to 80 TD updates, which amounts to 16 to 20 evaluation runs on the x-axis of the plot. This is a quite plausible and efficient result, as the best trajectories in the dataset need about 45 to 70 steps to reach the area of stability. Backpropagating the resulting, constantly low costs of these stable states to the initial states would need at least about as much TD updates plus / minus some generalization ability or error of the neural network.

In its second-last column, table \ref{tab:extended_evaluation_metrics} shows the performance metrics of the policy learned after $63 \times 4 = 252$ TD updates in run 1 in an extended evaluation from 5 random starting positions. Its values of $e_{\infty}=1.22$ is practically the same as the value of the best policy after 134 episodes, its time needed to stabilizing $N=75.6$ is only slightly longer. This is a general pattern; whereas many of the offline-learned policies achieve about the same performance in the balancing phase as the online-learned ones, they learn slightly less aggressive, and, thus, slower policies for swing-up. We could not reproduce the policy 134's slightly more aggressive swing up behavior in any of the offline runs, at least not reliably. Whereas the difference is within the variance bounds, thus, should not be over interpreted, this \textit{may} indicate that letting the policy really explore and let it "see" its own mistakes and wrong estimates of q-values, is somehow beneficial for a learning process, at least, where a single wrong action has significant consequences on the future costs and the data is still thin. Anyhow, where the impact is less grave, or, data density is higher, in the balancing phase, we can not see any differences between best online and offline policies.

During the flat part of the learning curves behavior is relatively stable and produces many good policies. Another policy, after $67 \times 4 = 268$ TD updates scores an $N=75.40$ and an $e_{\infty}=1.27$, for example.

But, a problem with a pure offline learning process is to determine, when its finished and when to actually stop it. Whereas the learning curves stay flat for a while (see fig. \ref{fig:learning_curve_real_cartpole_offline_log}), two of the three runs showed a clearly visible overfitting effect already within the first 100 episodes (400 TD updates), similar to the effect seen in the original interactive learning runs. The offline learning process needs to be stopped before running into this degrading effect.

As a result, it is not possible to run the offline learning process completely blindly, without any evaluation runs, especially on a new system, where learning parameters and resulting curves are not known beforehand. Evaluating every other iteration of the neural network, as we did here, is a good strategy for an empirical evaluation of the algorithm, but not a valid strategy for many practical applications, where reducing interactions with the system is the "true" optimization goal, especially of those where policies are bad and, thus, costs are high.

A good "passive" indicator for identifying the first candidates where evaluations should be done, for us, are the minimal, maximal and average q-values of the trained neural network. It needs several TD-updates until especially q-min and q-average spread, since as long as costs haven't been propagated backwards along a "goal-leading" trajectory, minimal costs and average costs appear to be very similar, due to no strategy for avoiding standard step costs consistently. This is a system-specific property, thus, needs some insight into the cost structure of an optimal policy, and might be different for different systems.

If the goal is to "recreate" a policy from the dataset (e.g. for testing different hyperparameters, changes in cost functions, etc.), thus, final values of q-min, q-max and q-average are already known from an initial successful run, waiting for the values of the offline training run to converge towards the known good values also is a good strategy for determining, when to start evaluation runs of good candidate policies. This strategy is task-independent, but needs knowledge from a previous successful run, or, as the method above, some theoretical insight into the cost structure and expected costs of an optimal policy.

\subsection{Offline Learning from the full dataset}
\label{sec:offline_full_dataset}

We repeated one offline run with the full dataset of 500 episodes from the original run 1 in order to find out, if it is possible to learn an even better policy than the one from the previous, now, on the full dataset with the increased number of transitions. It was possible to find policies with slightly better performance metrics, one of them (after 66 evaluation episodes, about 264 TD updates) with $e_{\infty}=1.199 \pm 0.082$ and $N=68.8 \pm 14.63$ (see last column of table \ref{tab:extended_evaluation_metrics}), the better average of $N$ likely not being significant, given the large variances of all tested policies. Another policy from an earlier iteration during the offline learning process (after just 22 evaluation episodes, about 88 TD updates) achieved a more stable, likely significantly better than policy 134, swing up behavior with $N=68.4 \pm 3.67$ but had a worse stability with $e_{\infty}=1.72 \pm 0.08$.

We note that better results may be possible with further hyperparameter tuning. In particular, we found the learning process with these larger datasets to be highly sensitive to the learning rate. For example, using the original default learning rate of $0.001$ resulted in less stable and repeatable learning processes and worse performing policies. Either reducing the learning rate, as we did here, or using larger batch sizes did help.

\subsection{Offline-Replay of the original training run}
\label{sec:offline_replay}

Finally, we wanted to get an estimate for how "lucky" we were with neural network initialization and data collection. The two sources of randomness in the otherwise completely deterministic NFQ algorithm are a) the collection of data during the exploration phase including the epsilon greedy choice of random actions as well as the intrinsic randomness of the physical system and the starting states, and b) the training of the neural network, which includes the random initialization of the neural network weights as well as the random sampling of mini batches in the batch-training. Freezing one source, the data collection, would allow us to have a look at the other factor and to find out, how dependent was the result on the randomness in the neural network initialization and training; would we always end up with the exact same policy given the same exploration and data?

In order to address this question, we repeated the offline training experiment, now not using the whole dataset as of after 134 episodes, but growing the batch of data from just one episode to 134 episodes, adding episode by episode after each iteration of the (four) TD updates. This can be understood as a "replay" of the original training run with exactly the same (random) exploration, with identical hyperparameters (learning rate set to 0.001 as in the original interactive learning run), but now with a neural network starting from a different random initialization.

The policy obtained after exactly 134 episodes of offline learning did show a less robust swing up behavior in the extended evaluation from 5 random starting positions with $N=104 \pm 45.96$ ($n=70 \pm 14.25$, $e_{\infty}=1.464 \pm 0.093$) but the policy just before that iteration after 133 episodes did show a good swing up behavior with $N=72.4 \pm 11.20$ ($n=67.8 \pm 8.70$, $e_{\infty}=1.396 \pm 0.108$) and also a good balancing performance of $e_{\infty}=1.396 \pm 0.108$. No single policy tested from the replayed dataset could match or beat the performance of the policy 134 from the original training run, in both, the swing up and balancing phase.

This indicates there is limited, but some relevant randomness in the random initialization and training of the neural network, as learning on exactly the same data and doing exactly the same amount of TD steps can lead to (slightly) different results just because of a different random initialization and sampled mini batches of the neural network. Again, this differently initialized network learning a different policy might suffer from not being able to see its own mistakes during an interactive learning process.

When letting the cloned policy 134 interact with the system starting from episode 135 onward, the performance in the swing up and balancing phase improves, eventually achieving a policy with a "repaired" swing up strategy and improved stability, with $N=75 \pm 11.6$, $e_{\infty}=1.338 \pm 0.054$).

This "replay" strategy is a third strategy to approach the stopping-problem of pure offline learning. The benefit of this strategy is, that parameters such as the learning rate, number of TD-updates and the stopping criterion do not need to be tuned for the offline learning process at all, as everything that worked initially during interactive learning can be left unchanged and used again for the offline learning process.

\subsection{"Reprogramming" or "hindsight relabeling" \cite{andrychowicz2017hindsight,li2020generalized}: bootstrapping policies to different cost functions from offline data}
\label{sec:reprogramming}

We can use these offline bootstrapping techniques not only to "clone" policies from another policy's data to the original task, but also to "reprogram" policies from the given data to different cost functions. As the full transition $(s_t, a_t, s_{t+1})$ is available, thus, the costs $c(s_t, a_t, s_{t+1})$ can be recomputed (relabeling the transition tuple) any time, we can use the same techniques as in the interactive learning process to change the cost function itself. This can be useful to, for example, refine the cost function as we go, e.g. narrowing the target margin of the pole position as the collected experience allows for learning more precise control.

Taking this idea to the extreme, we can also change the cost function completely, effectively switching to a completely different task and use case, learning a solution, a new policy, from the same, existing data in a purely offline manner.
\begin{figure*}[htb]
    \centering
    \includegraphics[width=\columnwidth]{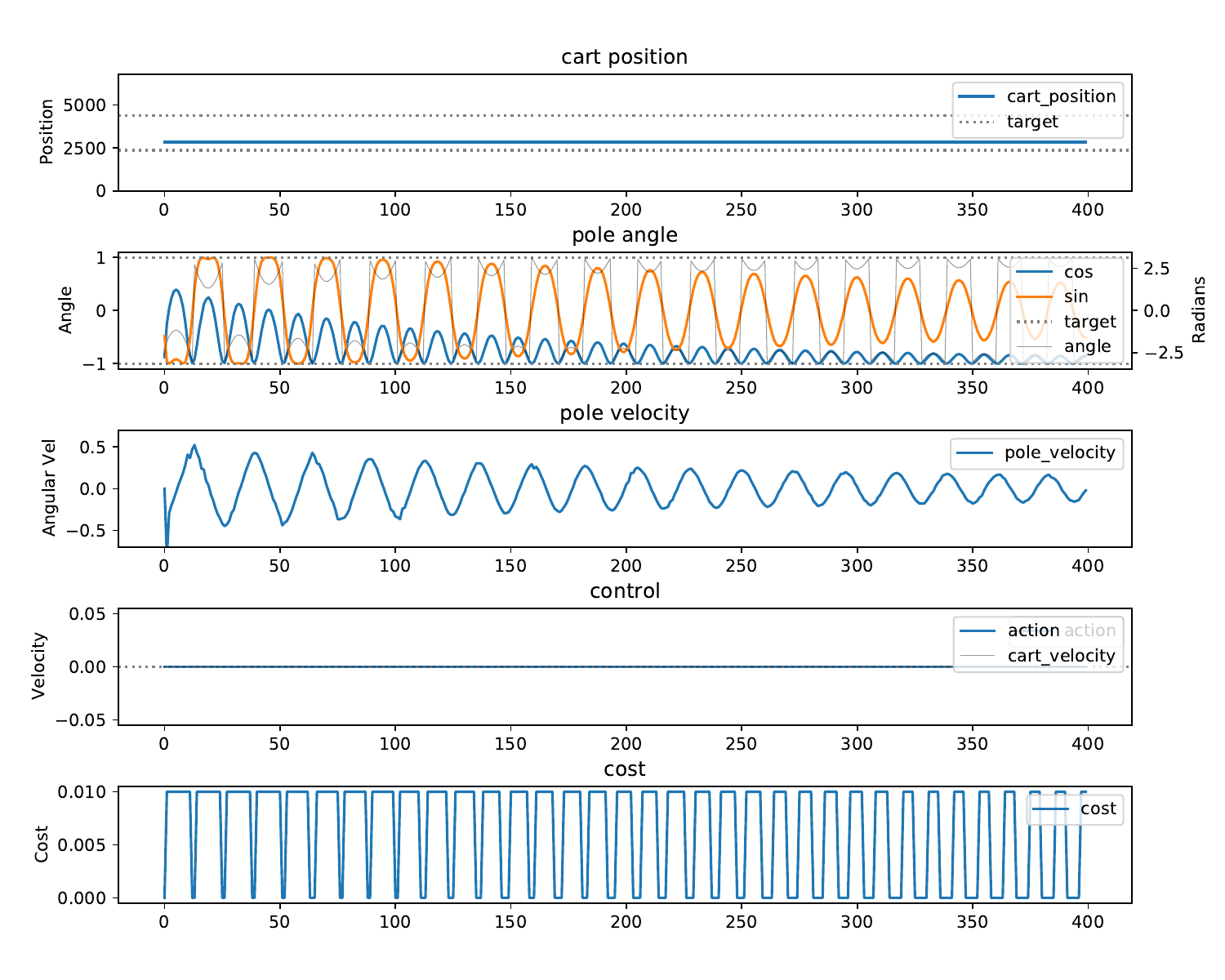}
    \includegraphics[width=\columnwidth]{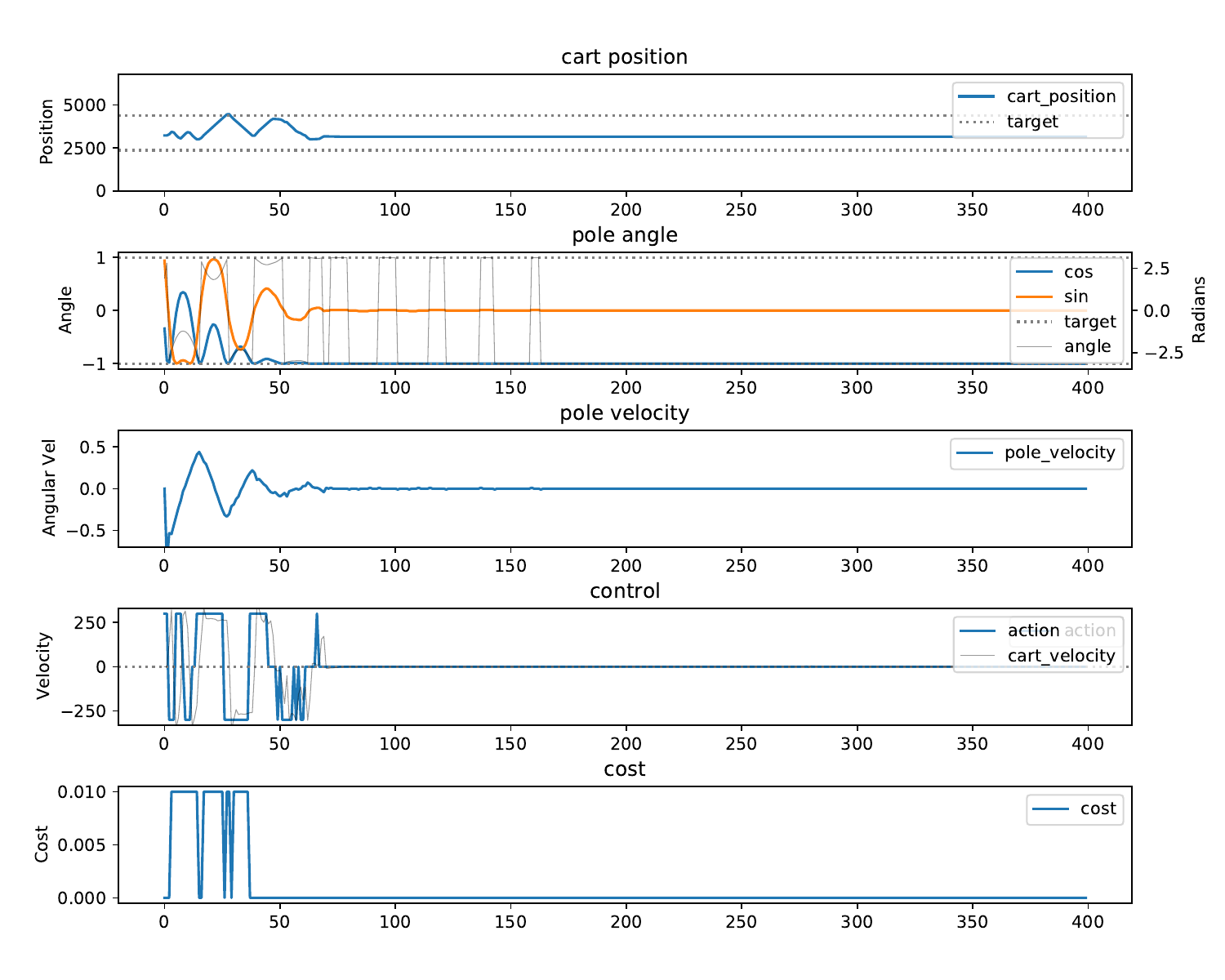}
    \caption{Sway killer policy trained purely offline on existing data (right) in comparison to free swinging behavior of the pole (left). The policy quickly removes the energy from the pole and brings it to a rest in the downward position with just a few oscillations.}
    \label{fig:sway_killer}
\end{figure*}

To demonstrate this, we define a new cost function $c(s_t, a_t, s_{t+1}) = c(s_{t+1})$ for a "sway killer" task on just the subsequent state $s_{t+1}$ of the transition, using only the cart position $x$ and the angle $\alpha$ of the pole of that subsequent state $s_{t+1} = (x, \Delta x, \alpha, \Delta \alpha)$:
\begin{equation*}
c(s_{t+1}) = \left\{
\begin{array}{ll}
1.0 & \text{if } x \in x_{\text{- -}} \\
0.1 & \text{if } x \in x_{\text{-}} \\
0 & \text{if } \frac{\cos(\alpha)+1}{2} \leq \theta_{\text{pole}} \text{ and } |x - \text{c}| \leq \theta_{\text{cart}} \\
0.01 & \text{otherwise}
\end{array}
\right.
\end{equation*}
where $x_{- -} $ and $x_{-}$ are the terminal region (highest cost, termination of episode) after hitting the endstop and the slightly wider soft stop region (high costs, episode continues) of the original cost function, $\theta_{\text{pole}}=0.05$ and $\theta_{\text{cart}}= \text{track-width} \times 0.15$ are the target tolerances on pole angle and cart position, respectively, and $c = \text{track-width} / 2$ is the center of the track.

The structure of the cost function is practically the opposite of the original cost function regarding the pole position. Low costs are now associated with the pole hanging down, forcing the controller to actively remove any energy from the system and to bring it straight down as quickly as possible. The cart position still needs to be near the center of the track, remaining unchanged.

We trained a new policy with this cost function in a purely offline manner, using only the data collected during the first 134 episodes of the original interactive learning run 5. All other parameters were kept the same as in the above learning experiments, this time leaving the learning rate at its default of $0.001$.

After 40 TD updates on the existing data, the learned policy demonstrated the behavior shown on the right of figure \ref{fig:sway_killer}. The controller was started from a random starting position with high energy on the pole\footnote{Whereas starting the CartPole from positions with high energy on the pole is simple in simulation, it takes a little effort on the real system. We use the well-known policy 134 to swing up and stabilize the pole in the upright position in a brief setup phase of 80 cycles / 4 seconds and then switch on the fly to the sway-killer policy we want to evaluate.} We can see the controller actively dampens the sway of the pole, taking away the energy of the pole and bringing it to a rest in the downward position with just a few oscillations. Afterwards, no further actions are taken, as constant zero costs are collected inside the reached target margins. A comparison to the free swinging behavior of the pole when no controller is active (see figure \ref{fig:sway_killer}, left) shows the great improvement that NFQ2.0 was able to learn from the data collected during a completely different task, not having actively interacted with the system for a single time.

An alternative to bootstrapping a completely new neural network from scratch is to use the existing policy as a starting point and "reprogram" it in several offline TD steps to a new cost function. This is especially useful, if the modification of the cost function is not too large and its general structure remains the same. Good examples for this would be decreasing a shaping component or by narrowing a target margin a little bit, as we will demonstrate in section \ref{sec:improvements}. When turning the use case and, with it, the optimal Q-function to its opposite, like we did above with the sway killer policy, limiting us to the reuse of the data and starting with a fresh neural network is more reliably leading to good results, fast.

\subsection{Summary of offline learning experiments}
\label{sec:offline_learning_summary}

The results of these offline learning experiments indicate, that learning and exploration can be separated from each other to a high degree, as described in \cite{pmlr-v164-riedmiller22a}, at least, if enough useful data is available and contains good trajectories reaching the goal state, as was the case here. If data coverage allows, switching the cost function to even completely different tasks is possible purely offline. But, at the same time, separating the learning process from the evaluation process is more difficult and poses the real problem, due to the inability of determining the best policy and right moment to stop the learning process blindly. Looking at the q-values and continuing with interactive learning and evaluation cycles after bootstrapping a new policy offline, either on the full data set or using the replay strategy, are valid strategies to at least reduce the number of necessary interactions with the system. A policy efficiently bootstrapped offline from existing data can be used as a good starting point for further interactive learning and evaluation.

We will make use of these techniques and learnings during later experiments, especially running the ablation experiments. Instead of running all experiments there from scratch, we might bootstrap neural networks with different hyperparameters on the available data, comparing the results of different offline-learning results, and, where useful, then continuing learning with them interacting with the real system. This helps us to create fair comparisons between different parameter settings while still bringing down the needed wall clock time for the empirical evaluations. We will also use some of these techniques to further improve the quality of the learned policies in section \ref{sec:improvements}.

\section{Ablation Studies}
\label{sec:ablations}

In the following we will investigate the influence of different hyperparameters and settings on the learning process and the resulting policies.

\subsection{To shape or not to shape}

In the evaluation, we used a shaped cost function, which included a term for the distance to the target angle of the pole. Such a shaping function is used to guide the policy towards a distant or hard to find goal area and to help the exploration and learning process. At the same time it might influence the quality of the learned policy in unwanted ways. For example, reaching some height quickly might produce lower average costs than a different policy that needs a little longer to reach the same height but ends up quicker in the tolerance area. A $\gamma < 1.0$ even increases the risk of this to happen, as the policy will prefer higher immediate rewards over longer term benefits to some extent.

Hence, if reaching the tolerance area (and stabilizing in it) as quickly as possible is the major goal, a simple, time optimal definition of the cost function without any shaping would be most appropriate. The time optimal formulation does not induce any wrong biases or unwanted "short cuts" to achieving low average costs other than reaching the goal area as quickly as possible.

We ran 5 independent learning runs as described in the evaluation chapter with the single change being the definition of the cost function:
\begin{equation*}
    c(s_{t+1}) = \left\{
    \begin{array}{ll}
    1.0 & \text{if } x \in x_{\text{- -}} \\
    0.05 & \text{if } x \in x_{\text{-}} \\
    0 & \text{if } 1-\frac{\cos(\alpha)+1}{2} \leq \theta_{\text{pole}} \text{ and } |x - \text{c}| < \theta_{\text{cart}} \\
    0.01 & \text{otherwise}
    \end{array}
    \right.
    \end{equation*}
    where the center of the track $c$, $x_{- -} $ and $x_{-}$ remain unchanged, $\theta_{\text{pole}}=0.3$ and $\theta_{\text{cart}}= \text{track-width} \times 0.15$ are the target tolerances on pole angle and cart position, respectively.

\begin{figure}[htbp]
    \centering
    \includegraphics[width=\columnwidth]{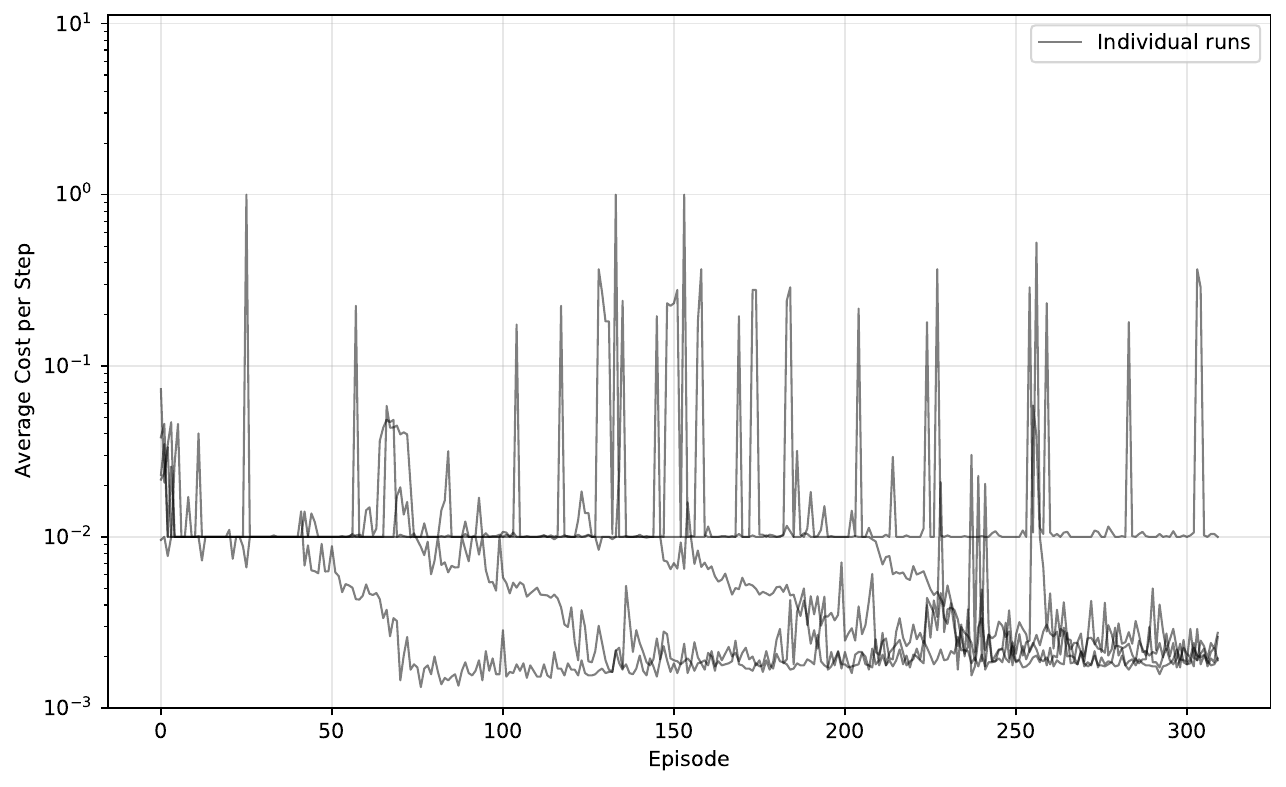}
    \caption{Learning performance without shaping rewards. The plot shows the average cost per step for 5 independent learning runs. Please note, that, whereas the shape and the variance of the learning curves can be compared to figure \ref{fig:performance_log}, the absolute values of the average costs cannot be directly compared, as we use a different cost function in this experiment.}
    \label{fig:performance_no_shaping}
\end{figure}
Figure \ref{fig:performance_no_shaping} shows the learning performance without shaping rewards. As can be seen, the learning process in average is much slower than with shaping rewards in figure \ref{fig:learning_curve_real_cartpole_offline_log}. The problem is that many episodes of exploration are somewhat wasted, until the agent eventually finds the goal area with the low costs and only then starts to improve the policy and to learn to swing up the pole. The moments of first detection can be  seen clearly where the curves of the individual runs "break" the average costs of 0.01 per step around episode 45, 70, 145, and 210. One of the runs didn't see the goal state once during all 300 episodes, and, thus, never learns any policy that gets below the default step cost of 0.01. Remembering that we had found proper policies by episode 70 when shaping rewards, this shows that the shaping rewards helped to guide the agent to the goal area and to speed up the learning process substantially. Also the high repeatability and low variance between individual runs that was observed in the shaping experiment is lost here.

Nevertheless, after seeing the goal state once, learning starts and the average costs drop quickly in a similar way as observed in the experiment with shaping rewards.

\begin{table*}[htb]
    \centering
    \begin{tabular}{l|c|c||c|c|c|c|c|c||c|c}
        &  \multicolumn{2}{c||}{\textbf{Shaped}} & \multicolumn{6}{c}{\textbf{Time Optimal}} & \multicolumn{2}{c}{\textbf{w. Demonstr.}} \\
        \hline
         & \textbf{Best} & \textbf{Avg} & \textbf{Avg} & \textbf{R1} & \textbf{R2} & \textbf{R3} & \textbf{R4} & \textbf{R5} & \textbf{D1} & \textbf{D1} \\
        \hline
        \hline
        Episode & 74 & $93.2 \pm 15.5$ & $161.5 \pm 70.6 $ & 134 & 202 & -- & 234 & 76 & 72 & 107 \\
        \hline
        $n$ & 59 & $65.6 \pm 6.58$ & $68 \pm 6.88$ & 72 & 73 & -- & 69 & 58 & 67 & 60 \\
        \hline
        $N$  &  59& $66.2 \pm 5.80$ & $68 \pm 6.88$ & 72 & 73 & -- & 69 & 58 & 67 & 60 \\
        \hline
        $e^{\infty}$ & 1.60 & $1.56 \pm 0.18$ & $2.20 \pm 0.92$ & 2.64 & 3.25 & -- & 1.24 & 1.65 & 1.58 & 1.48 \\
        \hline
        $e_T$ & 6.0 & $5.52 \pm 0.50$ & $6.60 \pm 1.55$ & 8.40 & 7.20 & -- & 4.80 & 6.00 & 5.40 & 4.80 \\
        \hline
        Successful & \checkmark & 5/5 & 4/5 & \checkmark & \checkmark  & - & \checkmark & \checkmark & \checkmark & \checkmark \\
    \end{tabular}
    \caption{Results of five independent training runs of NFQ2.0 with time optimal cost function (R1-R5), their average result as well as a comparison to the best (Best) and average result with the shaped cost function, copied from section \ref{sec:evaluation}. For each run, we show the metrics of the policy with the lowest average cost per step in their single evaluation run. Run R3 did not manage to swing-up the pole at all. The last two columns show the metrics of the best policies of two runs with the time optimal cost function that also received one episode which demonstrates reaching the goal area in the initial training data (D1-2). The time optimal cost function does not guide the exploration, thus, the best policy is usually found later (161.5 episodes in average), the progress is more random (higher variance with std=70.6) and sometimes even fails completely (R3). If the goal state is found, best learned policies perform similar, not better, regarding the swing up to the shaped cost function, and worse regarding the balancing phase. The later is not surprising, as the time optimal cost function does not have a cost gradient when within the goal area. The former indicates that the shaping did not harm--- at least not measurably--- and did not prevent learning a time-optimal strategy for swing up. Overall, using the shaped cost function is preferable and, in this setup, did not harm the learned policies performance. Results of D1 and D2 demonstrate adding one single successful demonstration is enough to overcome the time optimal cost function's disadvantage of not guiding the exploration and to bring learning speed and results to the same level as with the shaped cost function.}
    \label{tab:performance_metrics_time_optimal_scratch}
\end{table*}
Average metrics of the best policies are given in table \ref{tab:performance_metrics_time_optimal_scratch}. The best policies show a similar behavior in the swing up phase and slightly worse stability in the balancing phase as those found in the experiment with a shaped cost function (see sec. \ref{sec:evaluation}), when in both cases comparing those policies with the lowest average costs on their individual evaluation runs.

Given these results, in any practical or industrial application, where interaction time and robustness of the learning process are of concern, using the time optimal cost function directly is not advisable, at least not when learning from scratch.

Furthermore it should be noted that when comparing performance of different RL approaches it is very important to also compare the cost functions used. The same benchmark system becomes of a completely different difficulty, when using different cost functions, specifically, when shaping towards a goal area is involved.

\begin{figure}[htbp]
    \centering
    \includegraphics[width=\columnwidth]{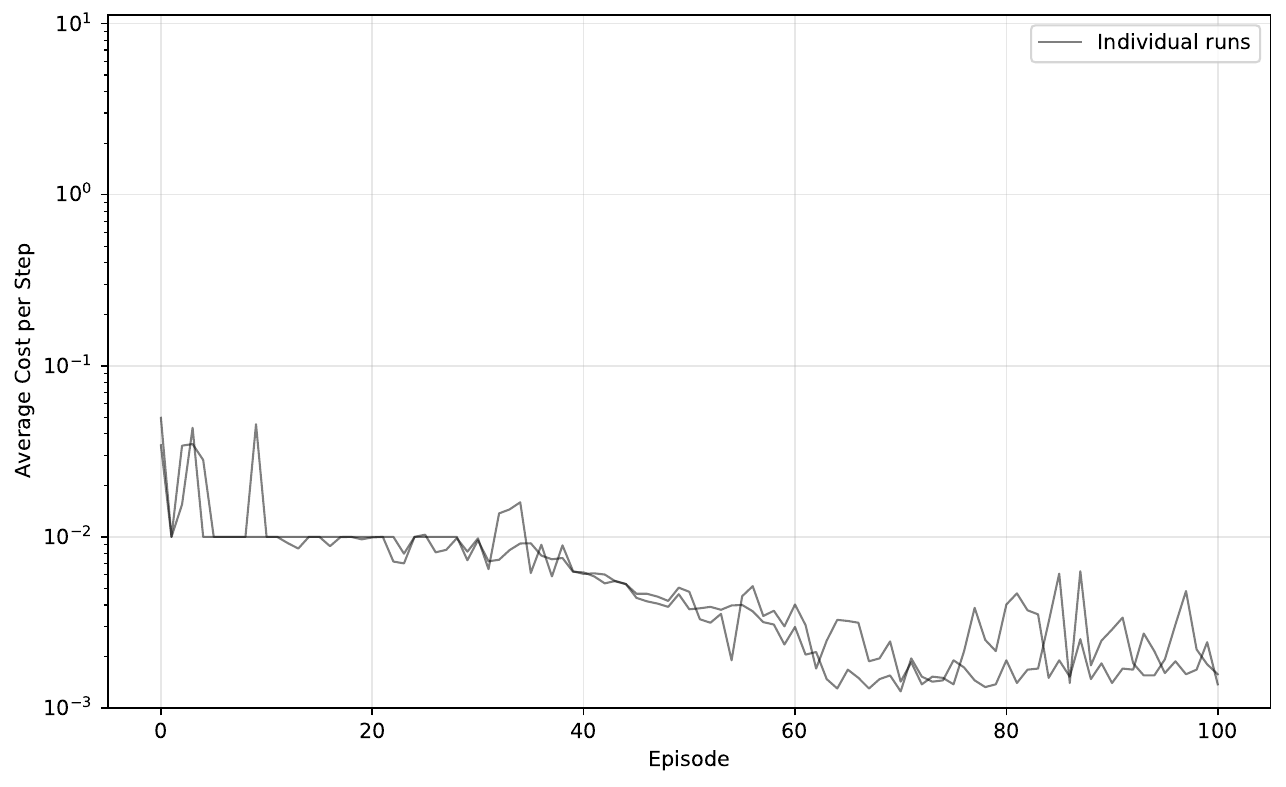}
    \caption{Learning performance without shaping rewards, as before in fig. \ref{fig:performance_no_shaping}, but this time with one single demonstration of reaching the goal area added to the initial training data. The single demonstration is enough to guide the learning process towards the goal area, to speedup the learning process and also to decrease the variance in its repetitions.}
    \label{fig:performance_no_shaping_with_demonstration}
\end{figure}
An alternative approach to work with the simple and time optimal cost formulation would be to somehow overcome the problem of not finding the goal area different than using reward shaping to guide the exploration to the goal area. Figure \ref{fig:performance_no_shaping_with_demonstration} shows two repetitions of learning runs with the time optimal cost function where we added one single "demonstration" of reaching the goal area by sampling a single random episode from a known good policy and adding this to the initial training data. Despite from having this single example, the setup and learning process was exactly the same in the previous experiments. As can be seen, this single demonstration is enough to speedup the learning process and also remove variance from its repetitions. Its very comparable to the curve of the shaped experiment. Average metrics of the best policies are also given in table \ref{tab:performance_metrics_time_optimal_scratch}. As can be seen, the results are significantly better than those without demonstration and match the results of the shaped experiment.

This result show clearly, how important it is to find the goal area at least once early in the learning process. It also shows, that the time optimal cost function is not inherently (a lot) harder to learn than a shaped cost function. Its main disadvantage is not guiding the exploration in the early process and, therefore, introducing a high variance into any evaluations of the learning process.

If a demonstration cannot be obtained from either an existing controller or a human expert, \cite{riedmiller2012tricks} contains several additional tricks to help guide the agent to the goal area without the need for shaping rewards or demonstrating proper trajectories, such as "the hint to goal" strategy where a few transitions or zero-q-targets within the goal area are added to the initial training data. Alternatively, the "reprogramming" technique we introduce in section \ref{sec:reprogramming} can be used to switch the cost function from one with cost shaping to a time optimal cost function on the fly, after having learned to reach the goal area, as we will demonstrate later.

\subsection{Actions in output layer}

Whereas the published network architecture of NFQ encoded the actions in the input layer, at the time, there were also early experiments with an output neuron per action, which was later popularized by the DQN algorithm \cite{mnih2013playing}. The DQN-like network architecture has the advantage of producing Q-values $Q(s,a)$ for all possible actions in a single forward pass, whereas the NFQ network architecture requires a separate forward pass for each action. NFQ has the advantage of being able to better generalize over the actions space, especially when the actions are not ordinal, but discretized values from a underlying continuous action space, like it is the case for the CartPole swing up problem. It's especially useful when the number of actions from the underlying continuous action space should be increased later on, as generalization among actions limits the necessary additional training data, as we will demonstrate later.

\begin{figure}[htbp]
    \centering
    \includegraphics[width=\columnwidth]{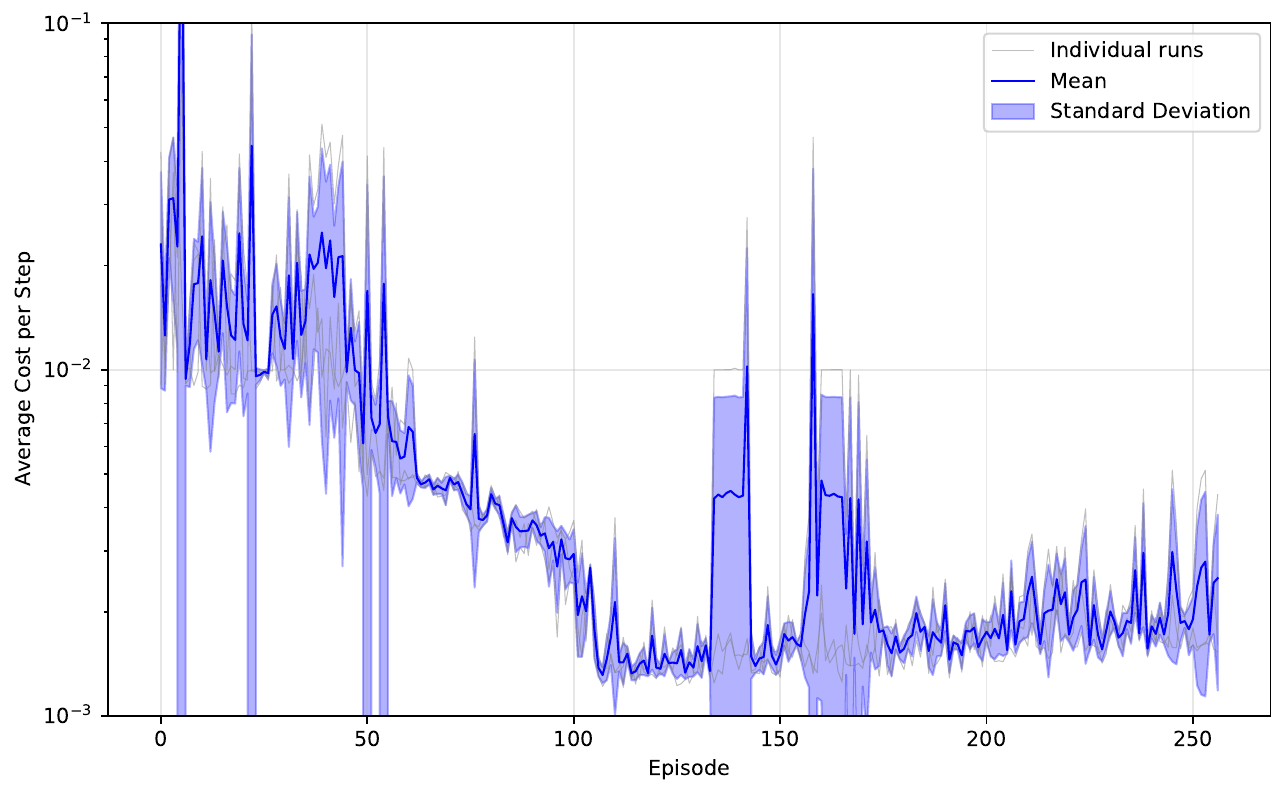}
    \caption{Learning performance with a DQN-like network, having an individual neuron for each action's Q-value in the output layer. The plot shows the average cost per step for 3 independent learning runs. Learning curves look very similar to the ones of the NFQ2.0 experiments with actions in the input layer. If anything, the DQN-like network seems to learn slightly slower, but this is not significant given the low number of repetitions.}
    \label{fig:dqnlike_performance}
\end{figure}
\begin{figure}[htbp]
    \centering
    \includegraphics[width=\columnwidth]{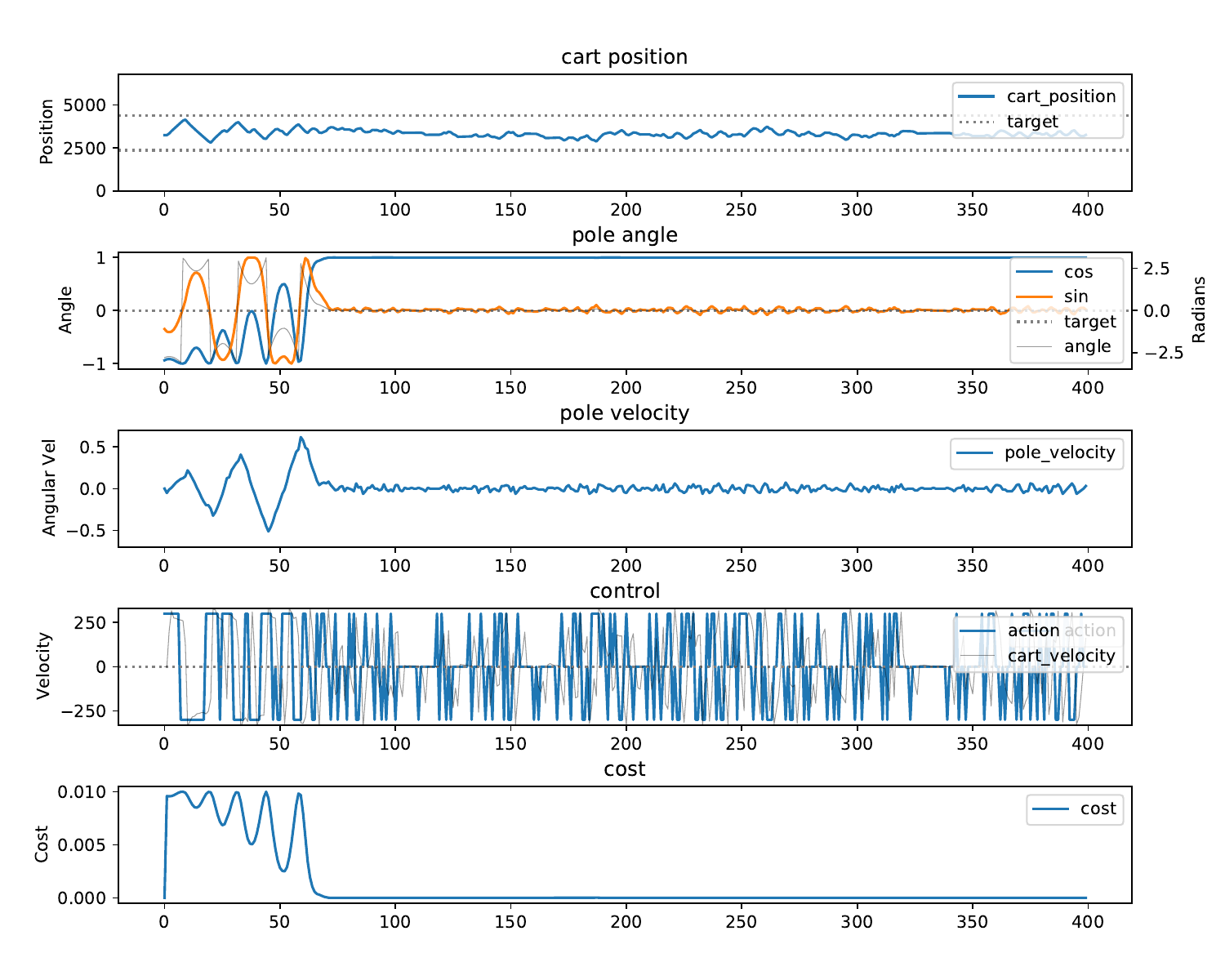}
    \caption{Evaluation run 112 of the best of the DQN-like network, having an individual neuron for each action's Q-value in the output layer. Its indistinguishable from policy 74 of the experiments with the action encoded in the input layer.}
    \label{fig:trajectory_dqnlike_run1_episode112}
\end{figure}

To address the somewhat open question, of whether or not the now-popular DQN-like network architecture also has some performance advantages over encoding the actions in the input layer, we ran another experiment with the same setup but with a DQN-like network architecture. The learning curves of three independent runs are shown in figure \ref{fig:dqnlike_performance}. As can be seen, the learning curves look very similar to the ones of the NFQ2.0 experiments with actions in the input layer. An exemplary evaluation episode is shown in figure \ref{fig:trajectory_dqnlike_run1_episode112}.

\begin{table}[htbp]
    \centering
    \begin{tabular}{l|c||c|c|c}
        DQN-like& \textbf{Avg} & \textbf{R1} & \textbf{R2} & \textbf{R3} \\
        \hline
        \hline
        Episode & $114.7 \pm 9.29$ & 112 & 125 & 107 \\
        \hline
        $n$ & $68.7 \pm 2.31$ & 70 & 70 & 66 \\
        \hline
        $N$  & $68.7 \pm 2.31$& 70 & 70 & 66 \\
        \hline
        $e^{\infty}$ & $1.6 \pm 0.31$ & 1.62 & 1.28 & 1.89 \\
        \hline
        $e_T$ & $5.4 \pm 0.60$ & 4.80 & -5.40 & 6.00 \\
        \hline
        Successful & 3/3 & \checkmark & \checkmark & \checkmark \\
    \end{tabular}
    \caption{Results of the best policies of three independent training runs (R1-R3) of NFQ2.0 with a DQN-like network, having an individual neuron for each action's Q-value in the output layer. First column shows the average metrics over the three runs. The quality of the policies is similar to the NFQ2.0 policies with actions in the input layer. If anything, the DQN-like network seems to learn slightly slower (best episode after 114.7 episodes in average versus after 93.2 episodes in average of the NFQ2.0 policies), but this is not significant given the low number of repetitions.}
    \label{tab:performance_metrics_dqnlike}
\end{table}
The metrics achieved by the best policies are also very similar to the ones of the NFQ2.0 policies with actions in the input layer (see table \ref{tab:performance_metrics_dqnlike}). If anything, the DQN-like network seems to learn slightly slower, but this is not a significant difference given the low number of repetitions. Thus, we cannot call a clear winner neither on the quality of the learned policies nor on the learning speed on basis of our experiments. But, definitely, the originally published NFQ network architecture is not inferior to the DQN-like network architecture.

Thus, the decision of either encoding the actions in the input layer or in the output layer is not critical for the learning process or the quality of the learned policies and depends more on other considerations, like possible limits on the inference speed or the ability to change actions during the application phase of the controller (as discussed later in section \ref{sec:improvements}).

\subsection{Network topology}

Riedmiller used to use a network with 2 hidden layers of 20 neurons each, all with a tanh activation function and the sigmoid for the single output neuron for the Q-value \cite{riedmiller2005neural,riedmiller2012tricks}. We ran two repetitions with this small topology, leaving all other parameters the same.

\begin{figure}[htbp]
    \centering
    \includegraphics[width=\columnwidth]{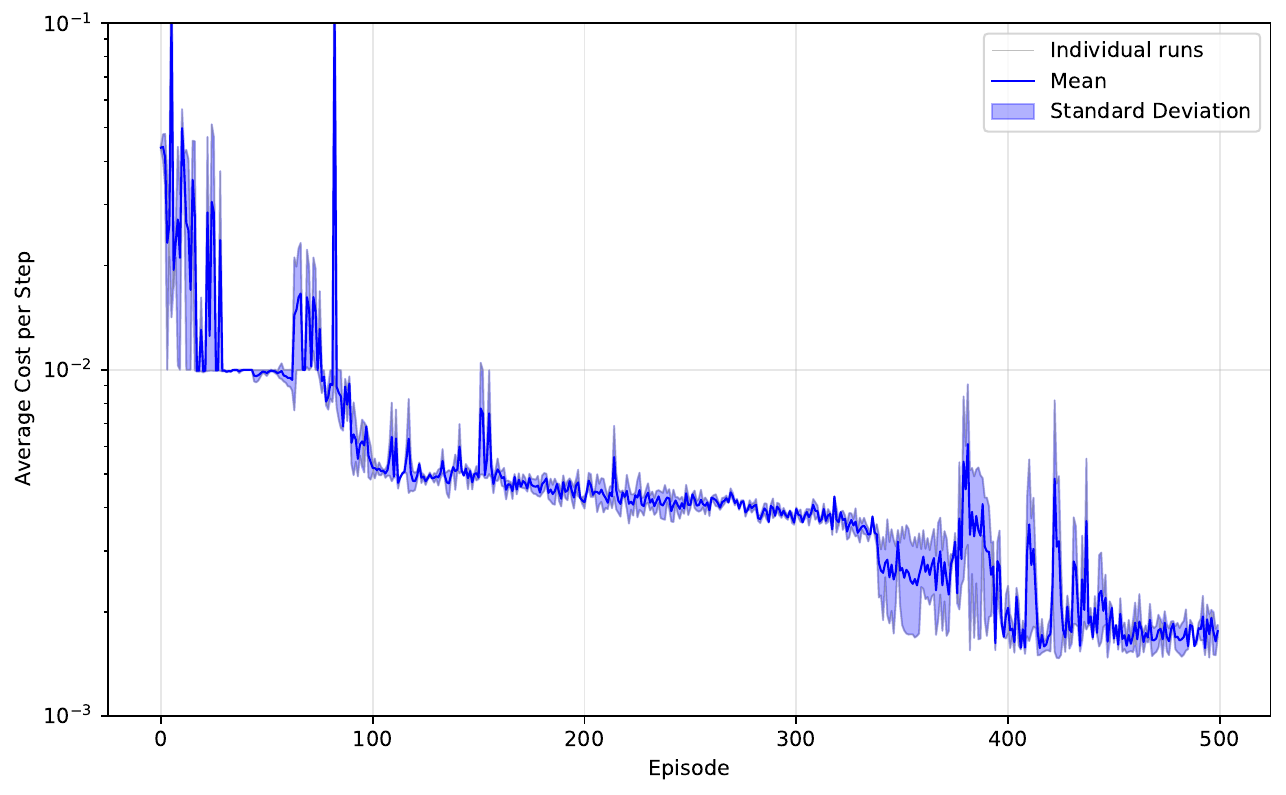}
    \caption{Learning curve of 2 independent runs with the small 20-20 network. The network learns significantly slower than the larger networks and might be hindering an effective exploration process. It got stuck in the windmill behavior for about 200 episodes, before rather suddenly jumping to a good balancing behavior within less than 10 episodes. Final performance is similar to the larger networks.}
    \label{fig:small_network_learning_curve}
\end{figure}
\begin{figure}[htbp]
    \centering
    \includegraphics[width=\columnwidth]{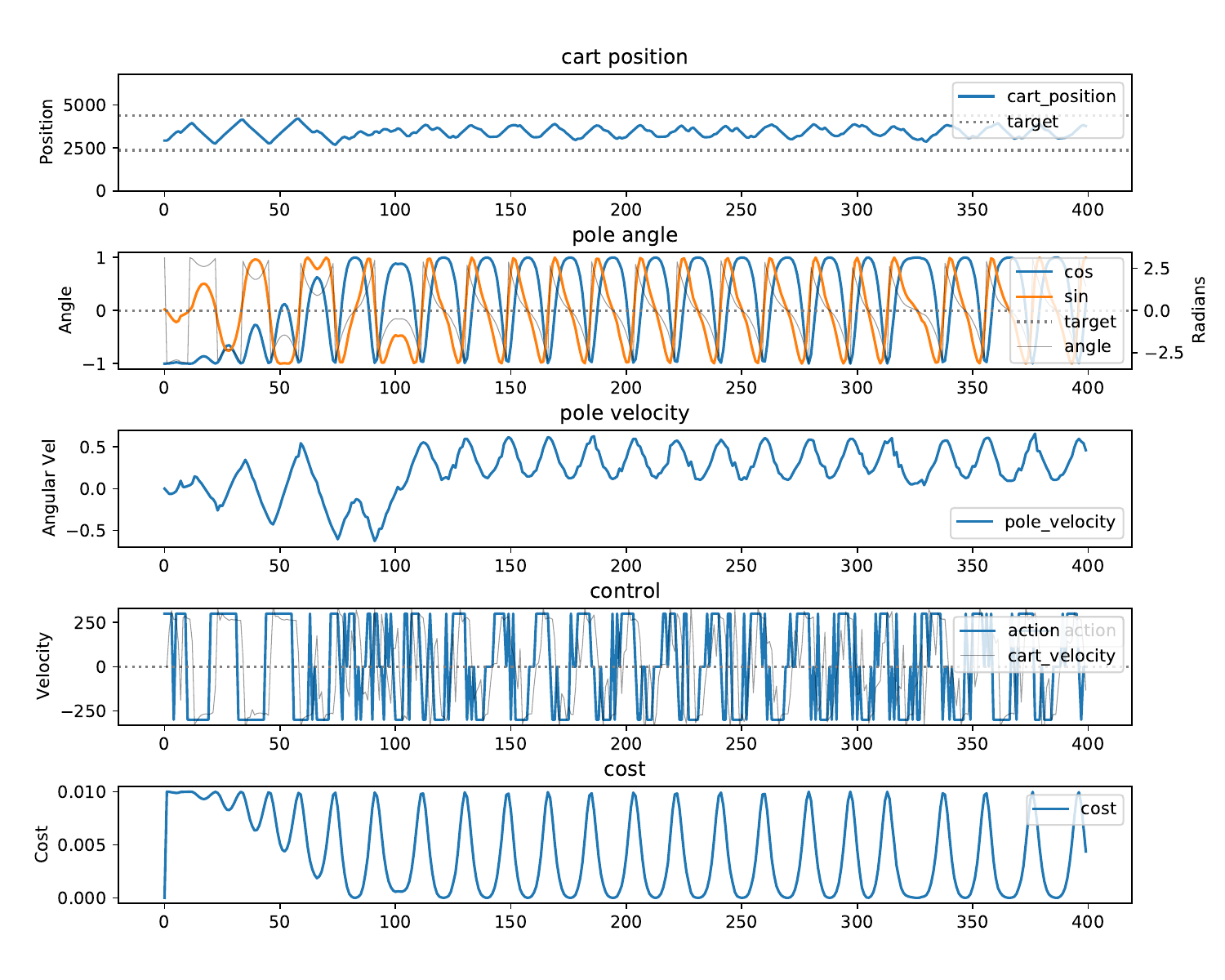}
    \caption{Exemplary trajectory of the windmill behavior that the small 20-20 network gets stuck in for about 200 episodes with only little improvement. The graph depicts evaluation episode 276. The windmill behavior is left for the very first time after episode 334.}
    \label{fig:trajectory_small_network_run1_episode276}
\end{figure}

The learning curve is shown in figure \ref{fig:small_network_learning_curve}. The small network learns significantly slower than the NFQ2.0 network and might be hindering the exploration process, because it needs too long to learn that increasing the energy leads to lower costs. It seems a little bit harder to train and is stuck for about 200 episodes in the windmill behavior, very slowly further improving the average costs, and, eventually, jumping to the balancing behavior from episode 334 (windmill) to 340 (balancing successfully for about 10s). An exemplary trajectory of the windmill behavior is shown in figure \ref{fig:trajectory_small_network_run1_episode276}.

\begin{figure}[bthp]
    \centering
    \includegraphics[width=\columnwidth]{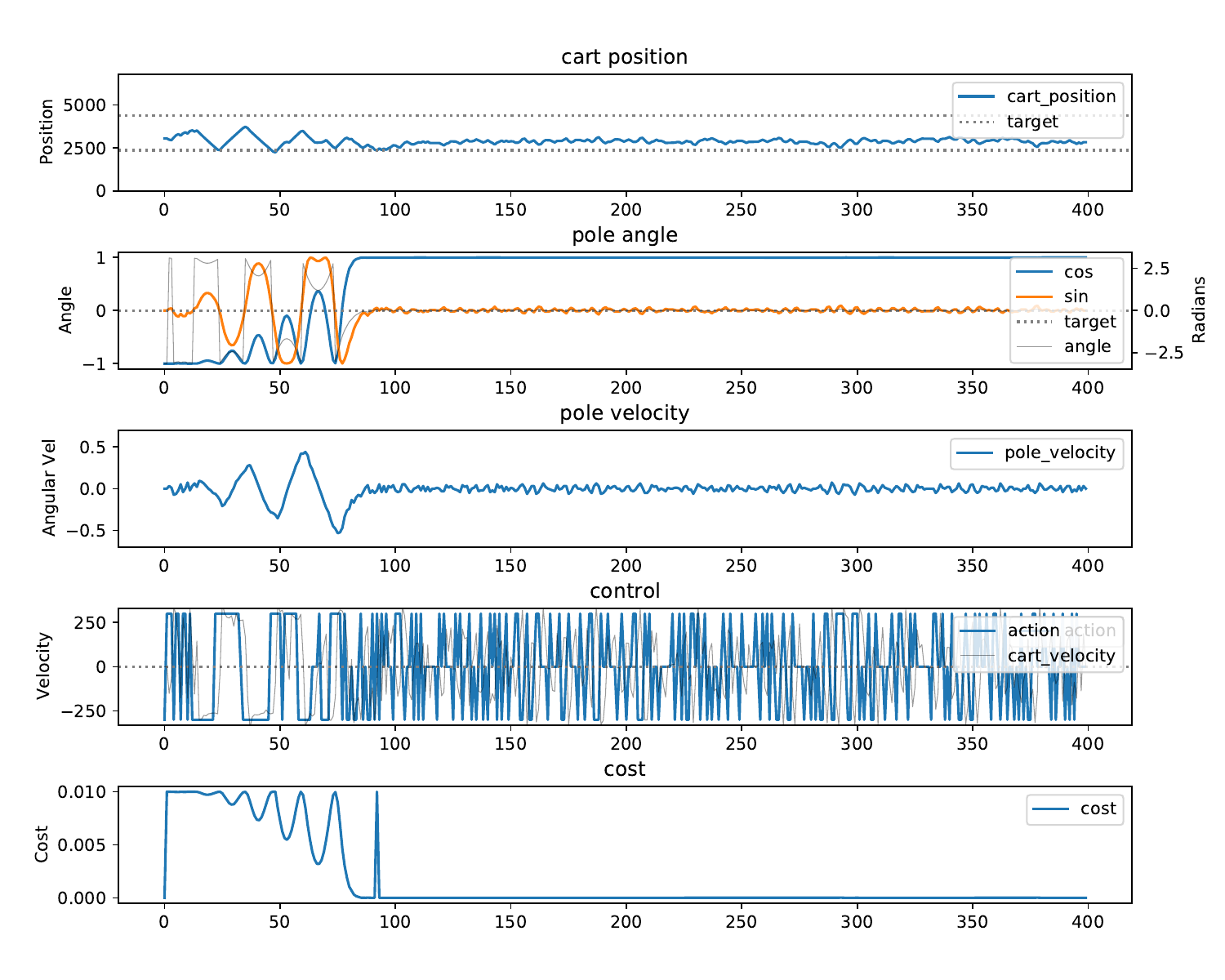}
    \caption{Exemplary trajectory of the best policy of the small 20-20 network. The graph depicts evaluation episode 472. The policy needs 85 steps to complete the swing up, loosing time especially in the beginning of the swing up, needing an additional half swing when compared to policy 74 using the larger network.}
    \label{fig:trajectory_small_network_run1_episode472}
\end{figure}

In run 1, the best policy was finally found after 472 episodes, with $n=N=85$, $e^{\infty}=1.335$ and $e_T=5.400$. In run 2, the best policy was found after 424 episodes, with $n=N=79$, $e^{\infty}=1.779$ and $e_T=s5.400$. The stability metrics are similar to the ones of the larger networks. The swing up behavior is similar, if not slightly slower, using an additional half swing (see figure \ref{fig:trajectory_small_network_run1_episode472}). It is worthy to note, that many policies show exactly the same swing up behavior, all needing 81 to 85 steps (run 1) and 78 to 80 (run 2) to complete the swing up. This behavior pattern seems more stable as in the training runs with the larger networks, unfortunately, also without any indication to further improve to less swings within the 500 episodes.

We repeated the experiment with the small network again in an offline setting, bootstrapping the network for 80 TD updates on the known-good dataset of policy 134, then letting it continue to actively explore, in order to find out, whether the quality of the learned swing-up behavior is limited by properties of the network topology itself or more indirectly by the slower exploration and learning process. The best network there managed to stabilize the pole within $N=69$ steps, several others achieved $N=75$ steps, with similar and even better stability metrics of $e^{\infty}=1.06$ for the very best evaluation attempt.

In summary, using a larger network topology that is easier to train (adapt), is clearly preferable as the small 20-20 network architecture does not necessarily worsen the policies performance, but is significantly harder to train, leads to a slower and more random exploration and overall learning process, probably with the risk of a full failure, if the windmill phase would have been left later. It also practically limits the policies' ability to learn an optimal, aggressive swing up behavior within reasonable time.

\subsection{Stacking}

In order to find out, what the influence of stacking on the learning process is, we ran several repeated experiments with smaller and larger stack sizes. Results are given in table \ref{tab:performance_metrics_stacking}. All policies were bootstrapped for 80 TD updates on the known-good dataset of policy 134, then let to actively explore further until no improvement was observed anymore, or 200 episodes were reached.
\begin{table}[htbp]
    \centering
    \begin{tabular}{l|c|c|c|c|c}
        Stack size& \textbf{1} & \textbf{2} & \textbf{4} & \textbf{6} & \textbf{12} \\
        \hline
        \hline
        $n$ & 45 & 68 & 71 & $71.4 \pm 10.3$ & 69 \\
        \hline
        $N$  & - & 68 & 71 & $71.4 \pm 10.3$ & 69 \\
        \hline
        $e^{\infty}$ & - & 1.39 & 1.26 & $1.24 \pm 0.06$ & 1.48 \\
        \hline
        $e_T$ & - & -4.80 & 4.80 & $5.76 \pm 0.29$ & 6.60 \\
        \hline
        Successful & -- & \checkmark & \checkmark & \checkmark & \checkmark \\
    \end{tabular}
    \caption{Results of the best policies learned with different stack sizes. All policies have been bootstrapped for 80 TD updates on the known-good dataset of policy 134, then let to actively explore further until no improvement was observed anymore, or 200 episodes were reached. All but $n=1$ achieved results that are similar to the default $n=6$ policy. The result of $n=6$ is the result of policy 134 copied from section \ref{sec:evaluation}.}
    \label{tab:performance_metrics_stacking}
\end{table}

As can be seen, there is a minimal stack size of $n=2$ that is needed to learn a proper policy that manages to stabilize the pole and keep it up the whole episode. We had the impression that q-min came down more quickly for the larger values of $n$ and observed, that proper policies were already found within the offline phase for $n=4$ and $n=12$. Further investigation with more repetitions would be needed to establish a significant result for this observation.

Our results indicate that a stack size shorter than the default of $n=6$ is good enough. But on the other hand, even the doubled stack size of 12 does not lead to worse results than the default. Thus, we can conclude that failing on the too large side nowadays is preferable, as too large stack sizes do not have a significant negative influence on quality of resulting policies, but too small stack sizes harm the quality of resulting policies or prevent learning a proper policy at all. This is clearly counter the intuition from the ``old days'', where a goal in experimental design was to keep the input dimension as small as possible by all means.

\subsection{Exploration}

\begin{figure}[htbp]
    \centering
    \includegraphics[width=\columnwidth]{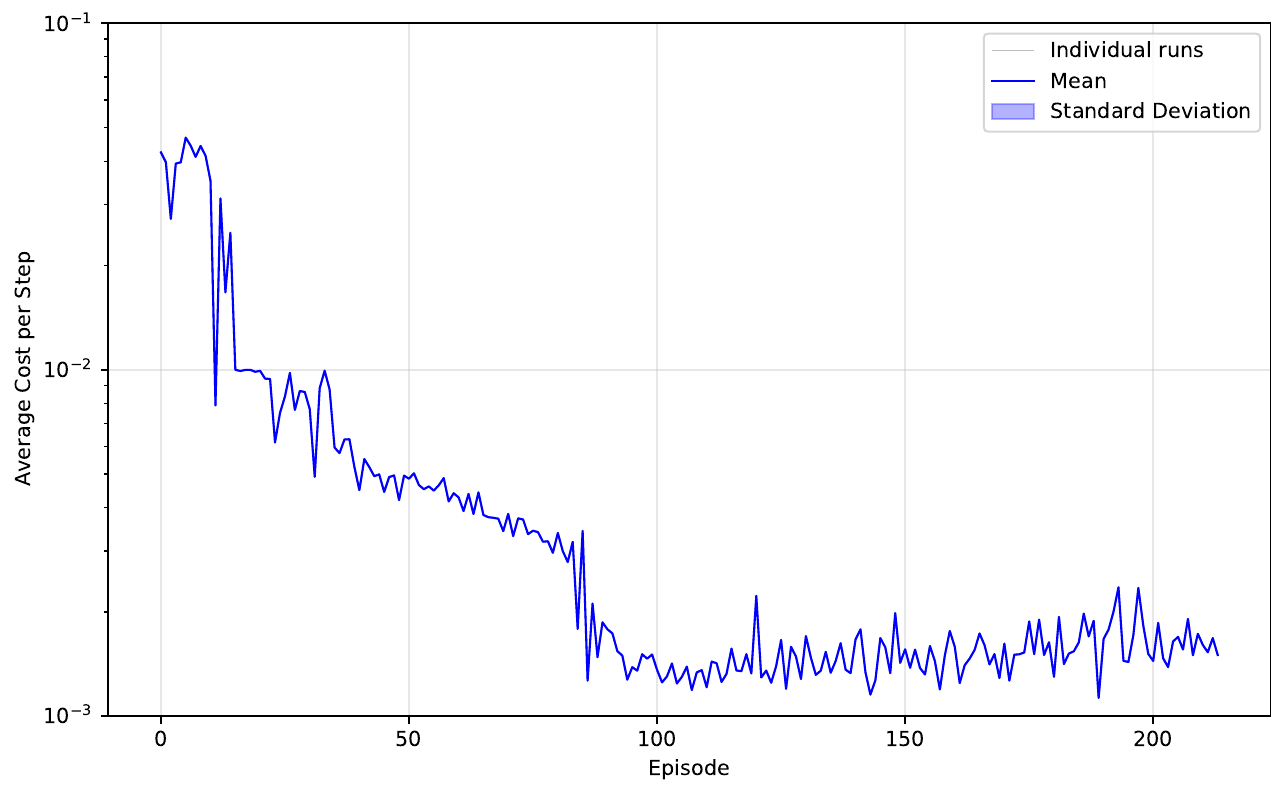}
    \caption{Learning curve of the experiment with $\epsilon=0$.}
    \label{fig:exploration_learning_curve}
\end{figure}

To test for the sensitivity of the learning process regarding the exploration rate, we ran several experiments with different constant values for $\epsilon$. We specifically tested $\epsilon=0.1$, and even $\epsilon=0$, as proposed by \cite{riedmiller2012tricks}.  Results are given in table \ref{tab:performance_metrics_exploration}.
\begin{table}[htbp]
    \centering
    \begin{tabular}{l|c|c}
        $\epsilon$ & \textbf{0} & \textbf{0.1} \\
        \hline
        \hline
        Episode & 189 & 93 \\
        \hline
        $n$ & 63 & 69 \\
        \hline
        $N$  & 63 & 69 \\
        \hline
        $e^{\infty}$ & 0.798 & 1.485 \\
        \hline
        $e_T$ & 4.20 & 6.00 \\
        \hline
    \end{tabular}
    \caption{Performance metrics of the best policies learned with different exploration rates. The best policy with $\epsilon=0$ was found after 189 episodes and exhibits a balancing behavior superior to all policies found in the 5 runs with the default setup with $e_{\infty}=0.798$ degree.}
    \label{tab:performance_metrics_exploration}
\end{table}
We could not find any significant differences in the speed of the learning process. Furthermore, both settings lead to good results that are comparable to the results of the default setup with one noteworthy exception, $\epsilon=0$ leads to by far the best performance in terms of stability metrics $e_{\infty}$ in the later phases of the learning process, with 9 policies achieving $e_{\infty}<1.0$ degree, and the two best achieving $e_{\infty}<0.8$ degree. It also seems that the learning process (see figure \ref{fig:exploration_learning_curve}) remains more stable and avoids overfitting a little longer than the runs with the default setup. Whether or not this result is significant or a lucky coincidence is unclear and would need further repetitions.

Nevertheless, the strategy of setting $\epsilon=0$ as proposed by Riedmiller is surprisingly still valid, also in our modernized setup and does not lead to worse results or a slower learning process, at least not, in the case where reward shaping is used. As this setting has the great advantage that explorative training runs could also be used as the evaluation run of the previous policy, the practical number of interactions with the system could be halved.

\subsection{Learning rate}

\begin{figure}[htbp]
    \centering
    \includegraphics[width=\columnwidth]{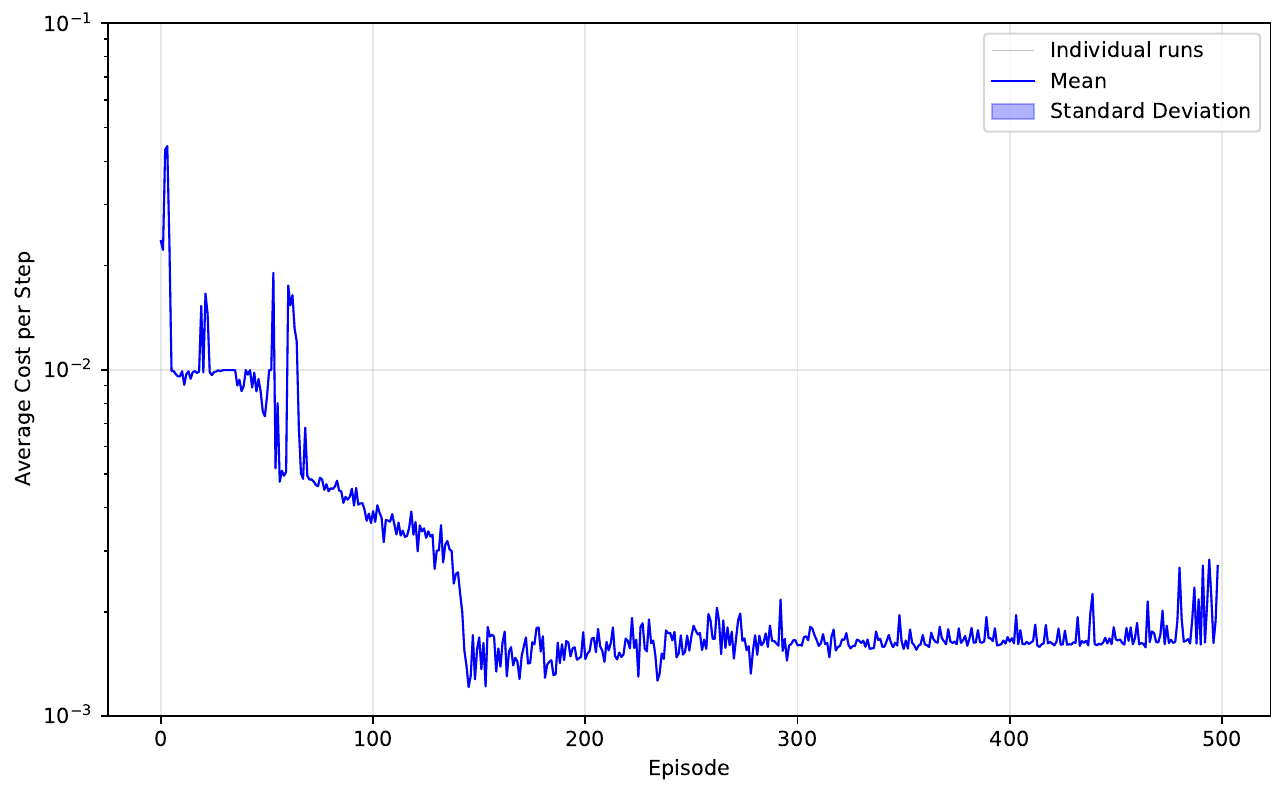}
    \caption{Learning curve of the experiment with a smaller learning rate of $0.0001$ instead of $0.001$.}
    \label{fig:learning_rate_experiment}
\end{figure}
We ran one experiment with a smaller learning rate of $0.0001$ instead of $0.001$  for 500 episodes. The learning curve is shown in figure \ref{fig:learning_rate_experiment}. This full repetition was slower than all 5 repetitions with the default learning rate presented in the evaluation section. It seems that the overfitting behavior after 200 episodes is less pronounced or at least delayed towards the end of the run, where variance in the policies' performance increased.

Most interestingly, there are many policies that achieved a more stable balancing behavior than all the other policies from the evaluation section with the default learning rate, with $e_{\infty}$ below 1.0 degree, the ten best all achieving $e_{\infty}$ below 0.9 degree. These 10 policies with the best stability metrics have been learned after e.g. 199, 426, 390, 498 episodes, further indicating longer progress in the learning process. At the same time, the swing up behavior of these policies was less fast than those of earlier policies.

The overall best policy according to the collected average step costs was found after episode 154 episodes, achieving $n=70$, $N=70$, $e_{\infty}=1.65$ and $e_T=-6.00$ which is comparable to the best policies of the default experiment, just found a bit later.

\subsection{Fresh Network each iteration}

\begin{figure}[htbp]
    \centering
    \includegraphics[width=\columnwidth]{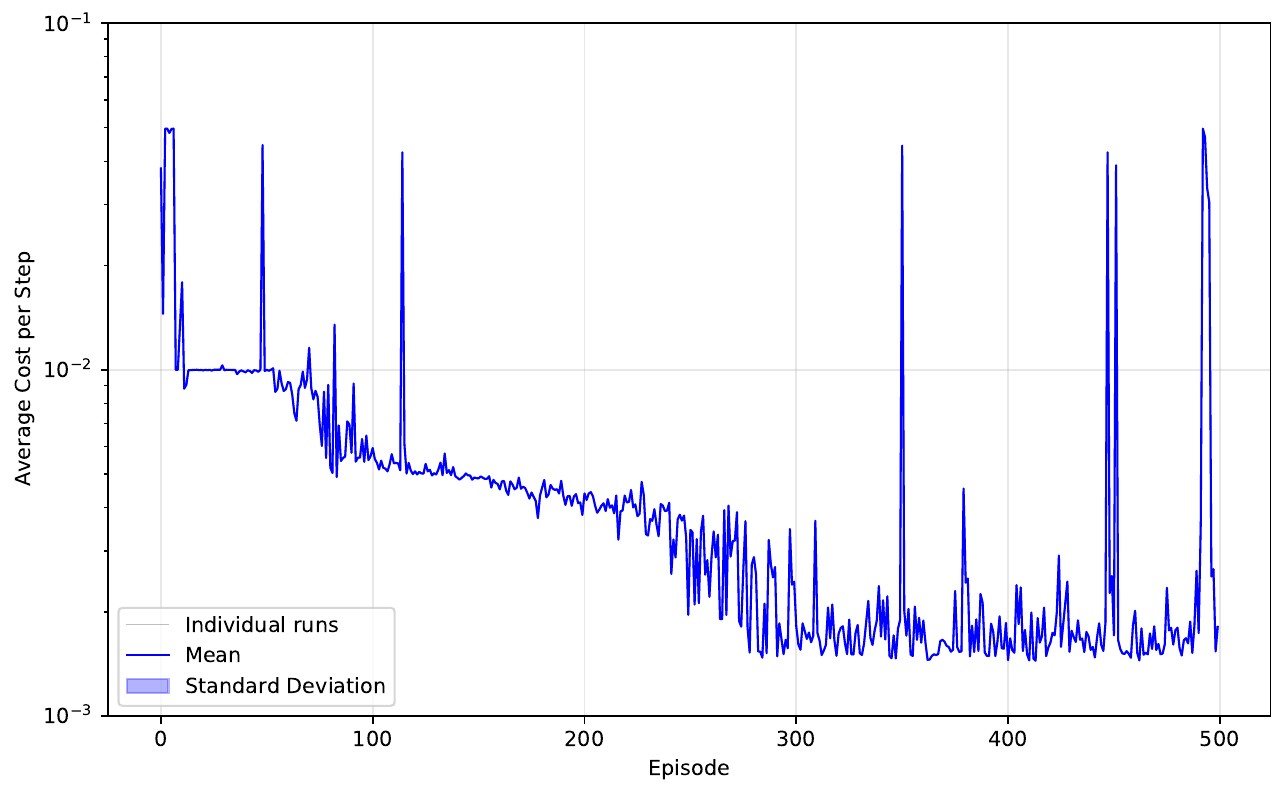}
    \caption{Learning curve of the experiment when using a fresh network and train it for 120 epochs after each episode. The agent needs significantly longer to learn to stabilize and balance and the variance from policy to policy is significantly higher, especially in the later phases of the learning process.}
    \label{fig:network_reset_learning_curve}
\end{figure}

We ran an experiment where we trained a fresh network from scratch for 120 epochs after each episode. The learning curve is shown in figure \ref{fig:network_reset_learning_curve}. The agent needs significantly longer to learn to stabilize and balance. This might be explained by only using 1 Temporal Difference step per episode, due to the long time needed to train the network from scratch to convergence. Considering number of epochs trained or wall clock time used, the process is clearly (by factors) slower. Furthermore, the variance from policy to policy is significantly higher, especially after the agent has learned to stabilize and balance. About every other policy in this phase fails to keep the pole up reliably, some not even manage to stabilize the pole once.

\begin{figure}[htbp]
    \centering
    \includegraphics[width=\columnwidth]{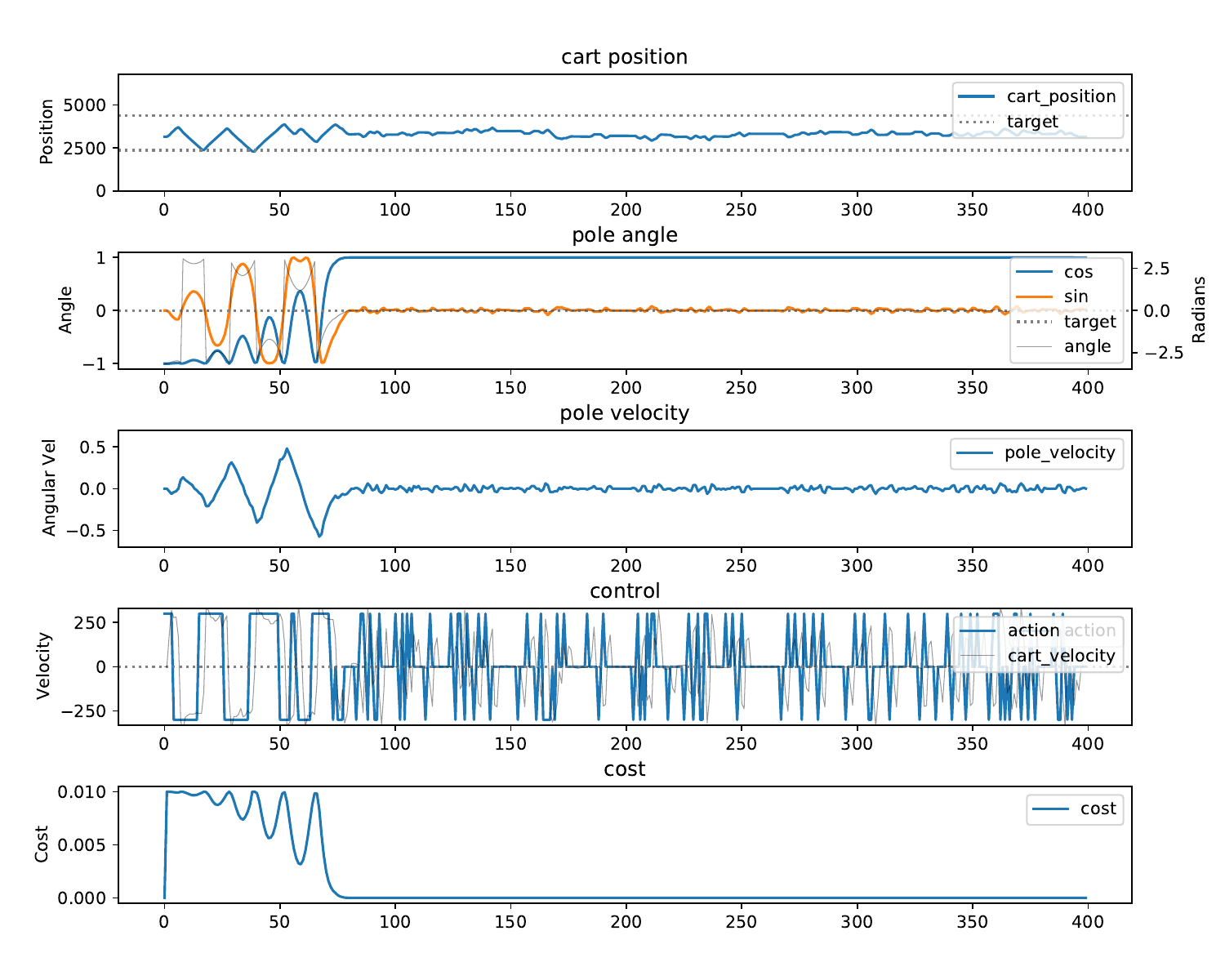}
    \caption{Exemplary trajectory of the best policy of the experiment when using a fresh network and train it for 120 epochs after each episode. The graph depicts evaluation episode 413. The policy's quality is comparable to the default setup, but the swing up behavior is less fast needing half a swing more than the best default policies.}
    \label{fig:network_reset_episode_413}
\end{figure}

The quality of the best learned policies is comparable to the default setup, with the overall best policy found after 413 episodes, achieving $n=N=77$, $e_{\infty}=1.182$ and $e_T=4.80$. An exemplary trajectory is shown in figure \ref{fig:network_reset_episode_413}.

\subsection{Normalization}

Riedmiller proposes to apply the normalization after each episode in \cite{riedmiller2012tricks}, as the errors the change of the scaling factors cause become negligible with time. This might work especially well when the neural network is trained from scratch after every single episode, but its unclear, if this also is negligible when continuing to train the same network.

We ran an experiment where we recalculated the normalization after each episode, as proposed by Riedmiller, and one additional experiment where we only normalized the input features up until episode 50 each 10th episode, stopping earlier as in the default settings.

We could not find any significant differences in neither the learning process' stability or speed, nor the quality of the resulting policies. Results of the best policies are the same as with the default settings, regarding both stabilizing speed and stability metrics. Network training seems to adapt to changed normalization factors with the supervised learning updates between the episodes. Thus, we can conclude that the re-normalization does not have a significant influence on the learning process, even when continuing to train the same network.

\subsection{Episode length}

Finally, we ran an experiment with half the episode length of only 200 steps, as proposed by Riedmiller in \cite{riedmiller2012tricks}. The best policy was found after episode 119, achieving $n=68$, $N=68$, $e_{\infty}=1.505$ and $e_T=6.60$, which is comparable to the best policies of the default experiment, just found a little bit later than the 93.2 episodes in average, and 108 episodes of the slowest run. The reduction in needed interactions (up to $119 \cdot 200 = 23\,800$ versus $93.2 \cdot 400 = 37\,280$) is significant, even when compared to the fastest run with the default settings (up to $74 \cdot 400 = 29\,600$).

\section{Further improving the quality of learned policies}
\label{sec:improvements}

In this section, we will use the tools and insights from the evaluation section and the ablation studies in order to demonstrate how to even further improve the policies we have learned so far in terms of the policies smoothness and performance measured by the metrics $N$ and $e_{\infty}$. Although we still continue working with the CartPole benchmark system, the tools and techniques demonstrated here easily translate to other systems, and, specifically well, to the engineering of solutions to industrial control problems.

The following description demonstrates an interactive approach to policy improvement, involving a single learning run with targeted manual interventions and adjustments and also a little (offline) backtracking. This mirrors how Deep RL would typically be applied by engineers in industrial brownfield settings, where an extensive grid search for good hyperparameters and multiple training runs from scratch are often impractical.

\subsection{Adding more actions to the controller}

The controllers so far had only two actions (bang-bang control) and a neutral action to chose from. Given this limited choice, the controllers can only become so and so smooth, effectively limiting the values of $e_{\infty}$ and $e_T$ that can be achieved physically.

NFQ2.0 encodes the actions in the input layer of the neural network. This has the advantage over encoding the action in the output layer of the network (as in DQN) that the network can more easily learn to use new additional actions, especially if they are "discretized" values from a continuous action space, thus, there is a relation between the different possible actions. As our action space is the target velocity of the cart, thus, "continuous"\footnote{The motor controller implemented in the PLC only allows 1024 different values, thus, is itself discretized. This is quite common in industrial control systems.}, we can take advantage of this by adding more actions to chose from.

For this, we start a new experiment with the well-known policy 134 from the evaluation section together with the data collected during the first 134 episodes of the original learning process. We will continue from there, effectively continuing the original learning process from episode 135 onward, but now "forking" it into a different direction.

First, we extend the set of actions to the values $(0, 1, 2, 3, 5, 10, 20, 30, 40, 50, 60, 90, 100, 150, 200, 300, 500)$ as well as their negative counterparts, resulting in 33 actions to chose from.

\begin{table*}[htb]
    \centering
    \begin{tabular}{l|c||c|c|c|c|c|c||c}
        \hline
          & \textbf{134} & \textbf{E.134} & \textbf{TOE.136} & \textbf{TOES.179} & \textbf{TOESP.224} & \textbf{TOESHP.280} & \textbf{TOESHPQ.341} & \textbf{DEL.341} \\
        \hline
        \hline
        Ep & 134  & 134 & 136 & 179 & 224 & 280 & 341 & 341 \\
        \hline
        $n$ & $71.4\pm10.31$ & $48.6\pm10.03$ & $49.4\pm5.54$ & $56.8\pm5.27$ & $55.8\pm10.11$ & $50.4\pm7.66$ & $42.6\pm4.13$  & $82.2\pm9.85$ \\
        \hline
        $N$ & $71.4\pm10.31$ & $64.2\pm15.16$ & $54.4\pm10.33$ & $70.2\pm27.90$ & $76.0\pm16.77$ & $50.4\pm7.66$ & $46.20\pm8.89$ & $82.2\pm9.85$ \\
        \hline
        $e^{\infty}$ & $1.24\pm0.059$ & $0.77\pm0.037$ & $0.85\pm0.027$ & $ 0.93\pm0.051$ & $0.84\pm0.032$ & $0.60\pm0.022$ & $0.54\pm0.024$ & $0.52\pm0.032$ \\
        \hline
        $e_T$ & $5.76\pm0.294$ & $3.24\pm0.612$ & $3.60\pm0.849$ & $3.12\pm0.449$ & $3.60\pm0.380$ & $1.92\pm0.240$ & $2.28\pm0.960$ & $3.24\pm1.804$ \\
        \hline
    \end{tabular}
    \caption{Performance metrics of different iteratively improved policies starting from the original policy 134. TOE: restart from episode 134 with time optimal cost function and extended actions, TOES: continuation, where only the target margin of the angle shaping is used, to create a cost gradient towards the center position, TOESP: continuation with and higher precision by shrinking the target angle margin, TOESHP: continuation with even higher precision by further shrinked target angle margin, TOESHPQ: continuation with an addition of a small cost to non-zero actions, with the goal of creating a quieter controller. All results in this table have been averaged over 5 longer evaluation runs.}
    \label{tab:performance_metrics_improved_policies}
\end{table*}

\begin{figure*}[htbp]
    \centering
    \begin{subfigure}[t]{0.48\textwidth}
        \includegraphics[width=\textwidth]{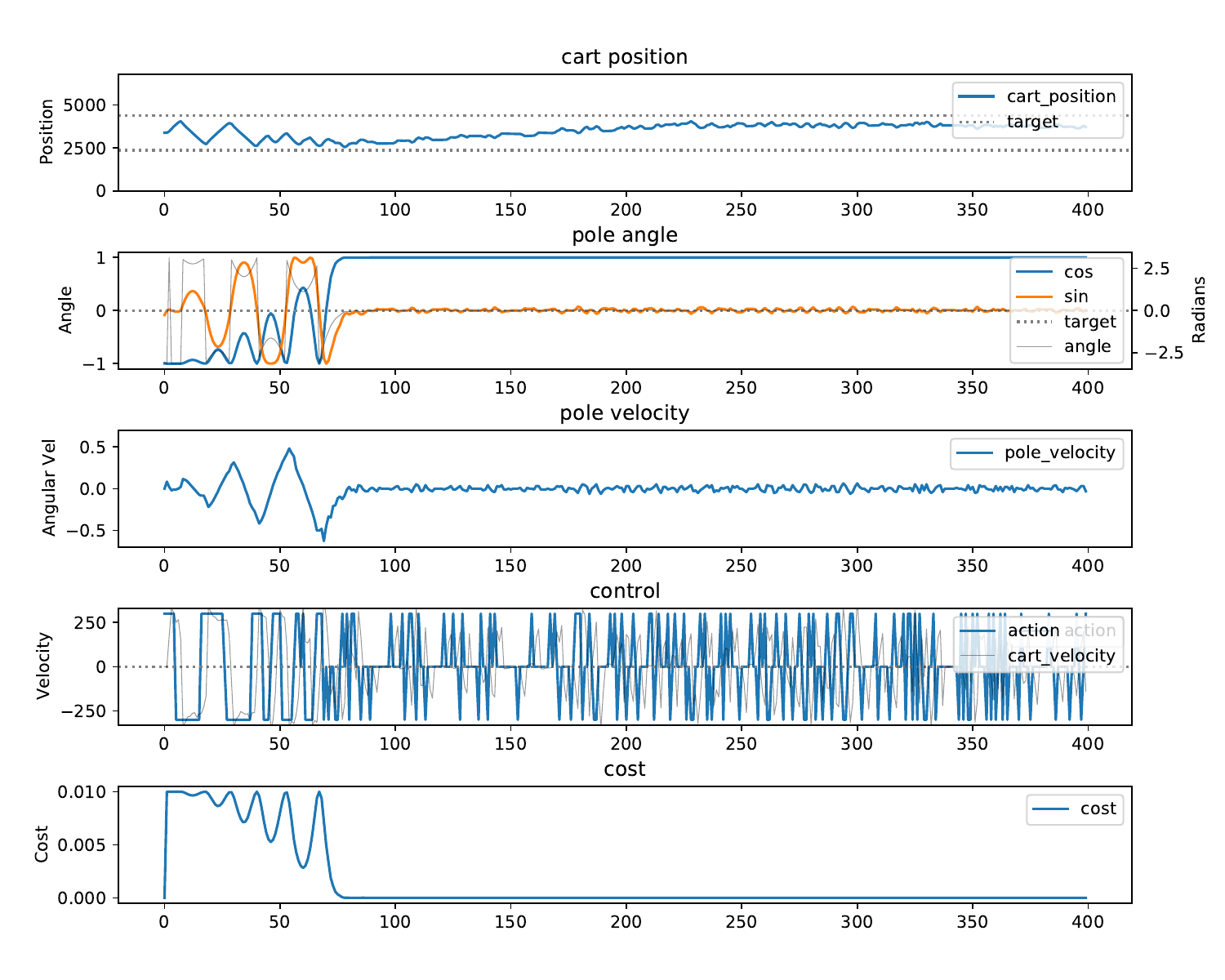}
        \caption{Evaluation of the original policy 134 from the evaluation section for comparison to the improved policies.}
    \end{subfigure}
    \hfill
    \begin{subfigure}[t]{0.48\textwidth}
        \includegraphics[width=\textwidth]{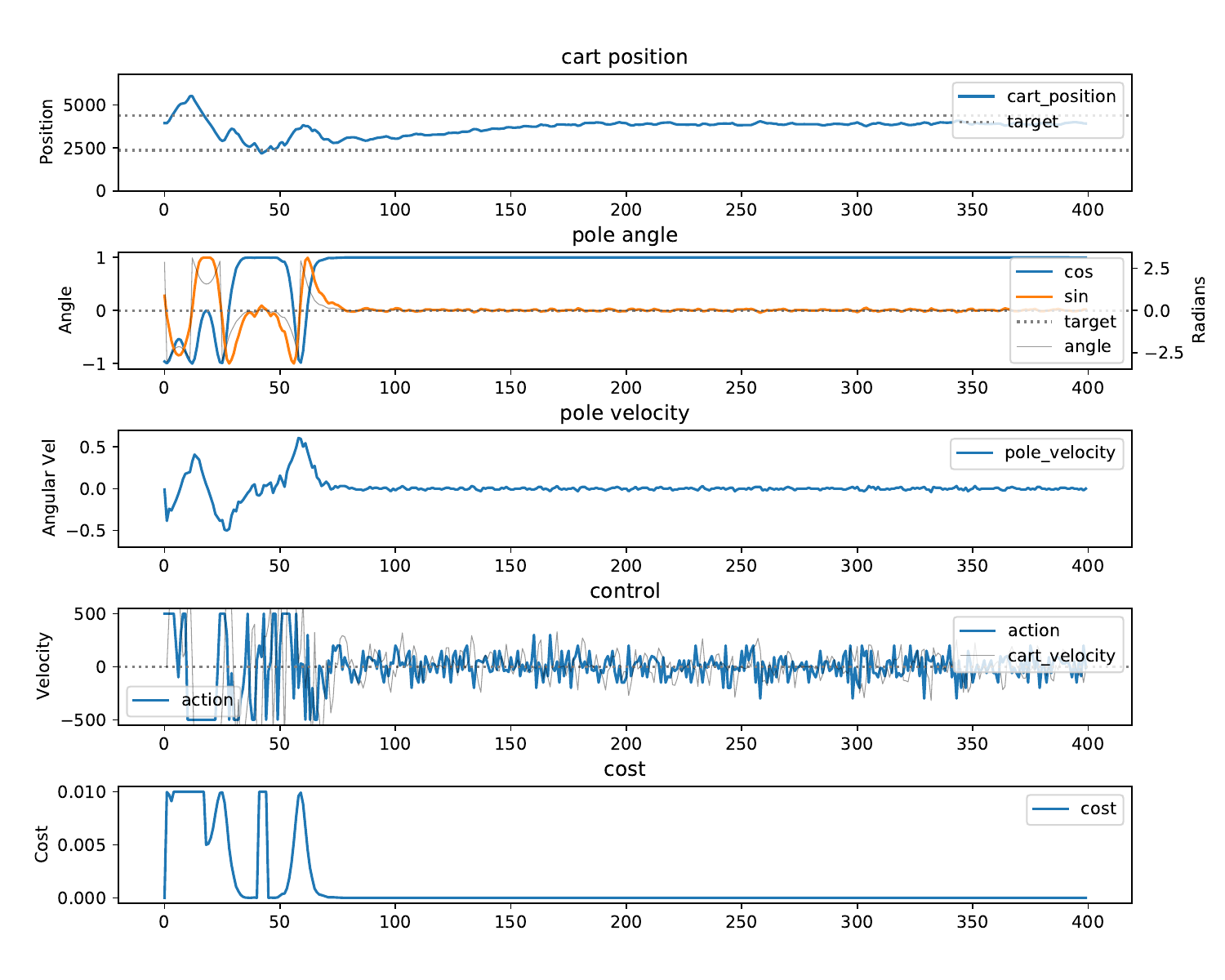}
        \caption{Evaluation of the policy 134 when extended to 33 actions, without any additional training just relying on the generalization
        abilities of the neural network. Although there are no transitions of the
        new actions in the training data at all, the extended policy makes use of the larger as well as the smaller actions in a useful way, achieving a more aggressive swing up--- at the cost of lower robustness--- and a slightly smoother trajectory with
        lower pole and cart velocities during the balancing phase. Note the imperfection during the initial swing up, where the cart leaves the goal area for a short time and the controller's counter reaction then letting the pole drop for a full additional swing.}
    \end{subfigure}

    \vspace{0.5cm}

    \begin{subfigure}[t]{0.48\textwidth}
        \includegraphics[width=\textwidth]{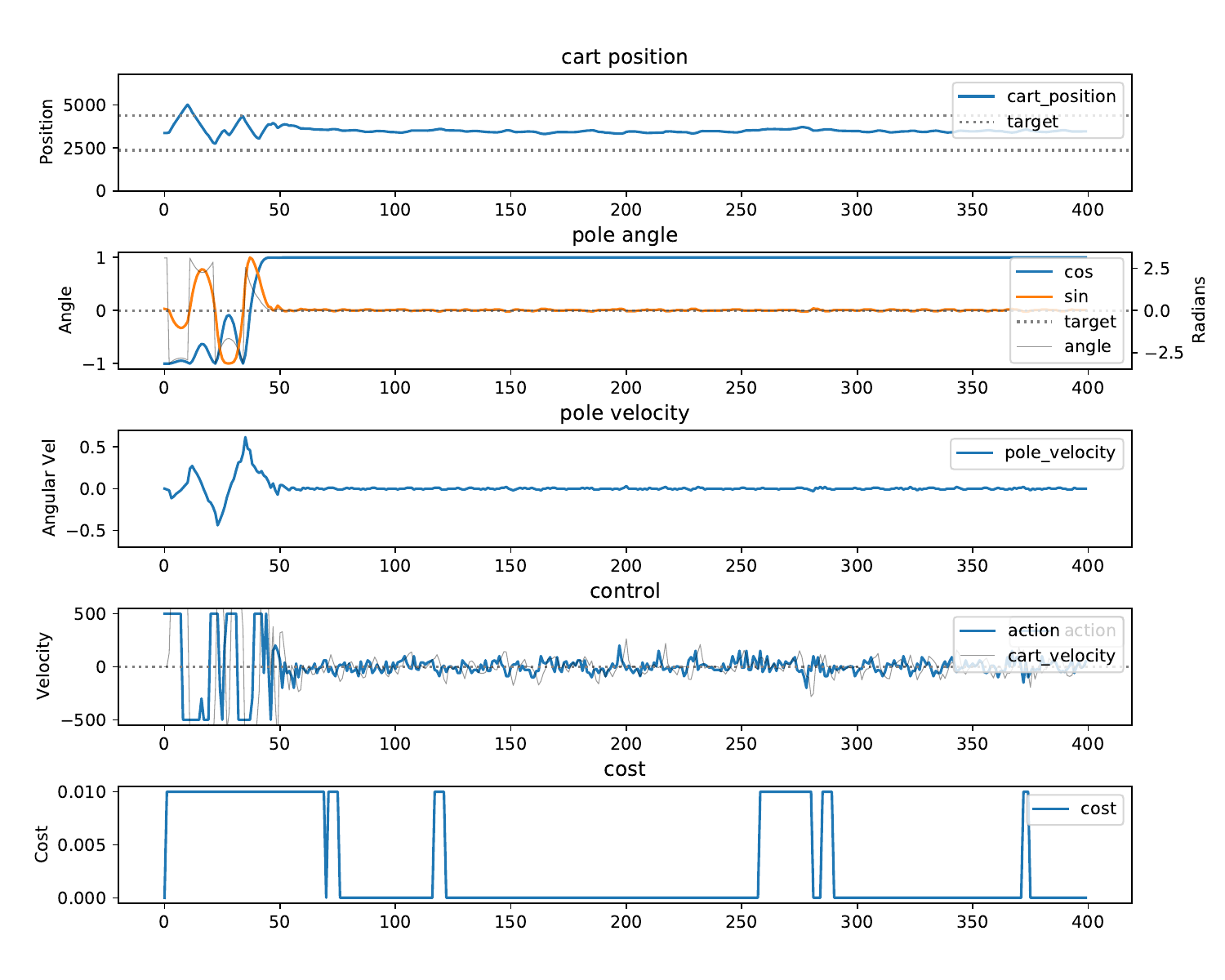}
        \caption{Evaluation of policy 280, which was trained to a higher balancing precision by shrinking the target angle margin to $\theta_{\alpha} = 0.0005$ and using a higher $gamma=0.999$. While the swing up behavior is visibly more aggressive and stability is also visibly improved, the bottom subplot shows that the controller sporadically leaves the narrow target margins, indicating some potential left for further improvement.}
    \end{subfigure}
    \hfill
    \begin{subfigure}[t]{0.48\textwidth}
        \includegraphics[width=\textwidth]{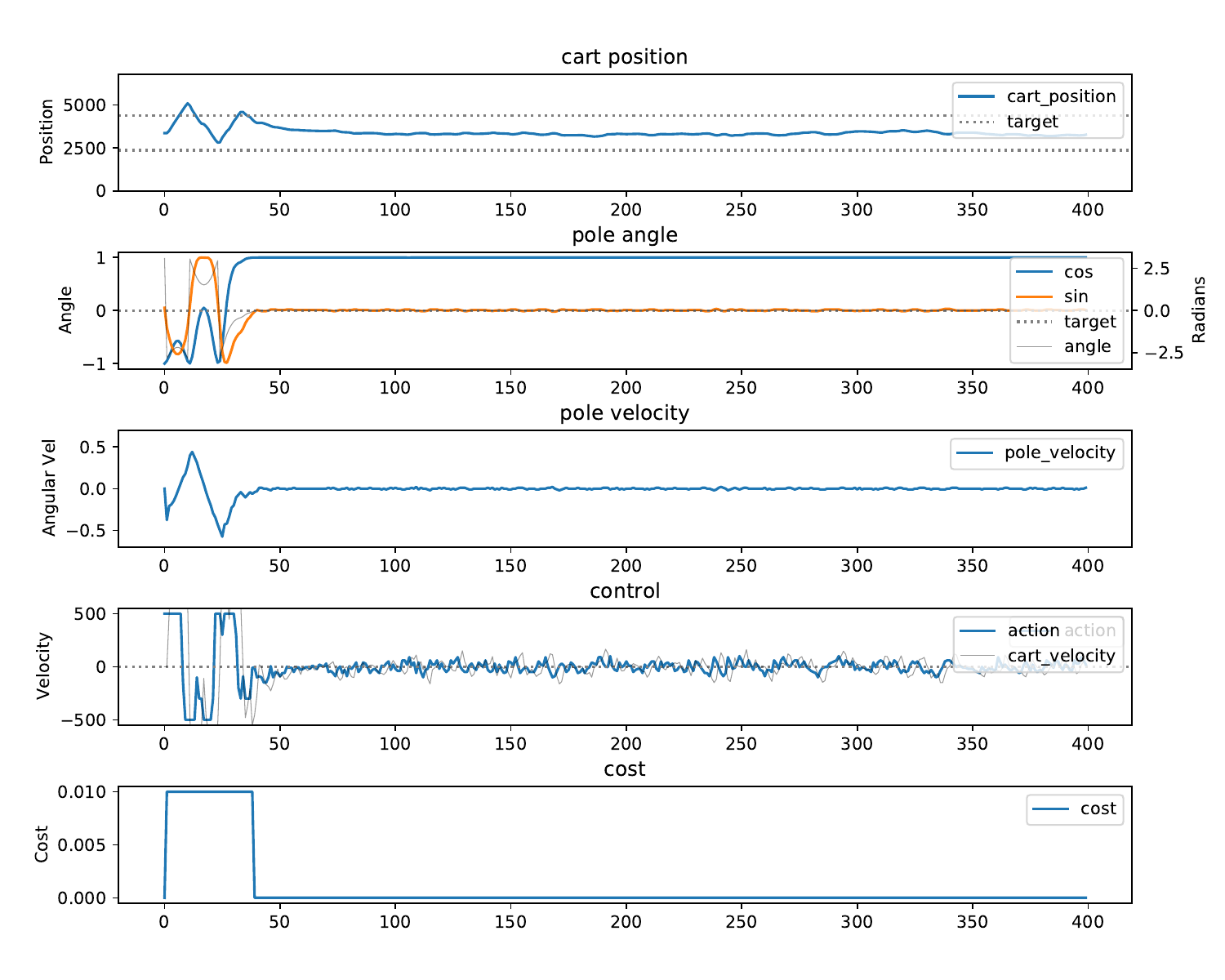}
        \caption{Evaluation of fast, reliable and energy efficient policy 341 which was trained with a high gamma value on a time optimal cost-function with narrow target angle margins, cost shaping within these margins towards upright position and a small extra cost on all non-zero actions. The policy shows a visibly more aggressive but nevertheless reliable swing up behavior and a smoother trajectory with lower pole velocities during the balancing phase. The chosen actions are significantly smaller than the ones chosen in (a) and (b) and (c).}
    \end{subfigure}

    \caption{Evaluation episodes of original policy 134 from the evaluation section (a), the same policy extended to 33 actions (b) an improved policy after 280 episodes (c) and the final best policy after 341 episodes (d). Please note, that the y-axis of the control diagram is scaled differently in some graphs, because of the new largest actions (b, c, d: 500, -500, a: 300, -300).}
    \label{fig:improving_policy_134_episodes}
\end{figure*}
It's important to note, that networks trained with NFQ usually immediately show some generalization abilities to additional actions "between" the previous ones, even if they have never been tried during exploration. To demonstrate this, we ran the neural network of the original policy 134 with the new actions an extended evaluation run first, before actually continuing the learning process. The behavior of a  repetition from a random starting state is shown in figure \ref{fig:improving_policy_134_episodes} (b). As can be seen, the network immediately uses not only smaller actions in between zero and the action it had explored (300, -300), but also larger actions (500, -500) in a useful way to swing up more aggressively. A quick evaluation of the metrics on 5 repetitions yields $N=64.2$ and $e_{\infty}=0.76$ (see table \ref{tab:performance_metrics_improved_policies}), which is already better than the original policy 134, without any additional exploration using the new actions.

\subsection{Switching to time optimal control}

We start the learning process with the extended action set and now also using a time optimal cost function. We switch from the shaped cost function to time optimal, in order to achieve the best possible swing up policy and to avoid the shaping function introducing any unwanted behavior collecting "early" low costs already in the swing up phase.

The adaptation of the estimates of the q-values to the changed cost function will take several TD update steps but generally is not a problem, if the change is not too drastic and the neural networks have not been trained too long already. As said before, we can continue to use all the sampled transition data in each iteration of NFQ2.0, as we can recalculate the costs $c(s,a,s')$ for each stored transition on the fly, whenever we change the cost function.

Already the very first iterations show a similarly aggressive, but visibly and measurably more robust swing up behavior in the evaluation episodes (see table \ref{tab:performance_metrics_improved_policies}, policy TOE.136). But continuing the learning process further, we notice the stability $e_{\infty}$ and $e_T$ to degrade slowly but continually.

From our experience its usually the case that the policy degrades a little bit during the first iterations before it comes better again after adding additional actions or changing the cost function. Estimates of the q-values need to be updated and propagated backwards along the trajectories.

But, in this case, there is also a potential problem in our change: with taking away the shaping term penalizing small values of $cos(\alpha)$ we also took away any pressure to optimize the pole angle further when within the target angle margin. As the policies never leave this margin in the balancing phase, although their deviation increased slightly, there is no pressure to choose those actions which would keep the pole more upright over the other actions, that don't leave the margin but won't reduce the deviation.

\subsection{Shaping within the target angle margin only}

To correct this, after episode 143, we do a little back tracking and change the cost function again, this time adding the shaping term back, but only within the target angle margin. Like the time optimal cost function, this cost function only gives a cost below the standard costs after reaching the target angle and cart position margins, and the standard costs outside this region. But, at the same time, it creates a very small but non-zero cost gradient towards the center position again within the target area. At the same time, we reduce the learning rate by one order of magnitude to 0.00001, slowing the learning process, but also avoiding the overfitting we have seen in the original evaluation runs and allowing the neural network to also adapt to the significantly lower cost gradient in the target area.

We continue the learning process from episode 143 starting with 40 offline TD steps on the existing data, to let the network quickly adapt to the new cost function and repair the degradation of the stability we caused with the removal of the shaping term before continuing the interactive learning process.

Please note, that we now can use the offline TD steps for bootstrapping the new cost function in a meaningful way without any additional data collection, as we have collected about 4000 transitions with the new, additional actions meanwhile. In this sense, the 10 explorative iterations of NFQ2.0 with the "wrong" cost function are not a waste of time, because we explored more necessary data and can "repair" the wrong observed costs with a few offline reprogramming steps while recalculating the transition costs with the corrected cost function on the fly. No precious data is lost, all data is used.

Continuing the learning process further, we can see the degradation of the stability to be undone and the policy to become smoother again in the balancing phase but still not reaching the $e_{\infty}$ of the original shaped policy 134.33 (for metrics of an exemplary policy from this phase see table \ref{tab:performance_metrics_improved_policies}, policy TOES.179).

\subsection{Increasing precision}

Therefore, we further increase the precision rewarded by the cost function, by reducing the target angle margin in two steps of an order of magnitude each, first from $0.3$ to $0.05$ after 200 episodes and then, after watching the controller's behavior, especially its ability to stay within the target area and to collect near-zero costs for a while, to $0.0005$ after 231 episodes. With the first precision increase, we also increased the gamma value to $0.999$ from $0.98$ for increasing the time horizon. An intermediate result after 224 episodes (after step 1, before step 2) is listed in table \ref{tab:performance_metrics_improved_policies} as TOESHP.224.

We let the learning process continue until we do not see any improvements for a while. The resulting policy is listed in table \ref{tab:performance_metrics_improved_policies} as TOESHPQ.280. With a stability of $e_{\infty}=0.60$ and maximum deviation of $e_T=1.92$ it is the most stable policy so far, clearly outperforming all previous policies including the reward-shaped policies. At the same time the policy exhibits a very aggressive, fast and at the same time robust swing up behavior with $N=50.4$ and $n=42.6$, also clearly outperforming all previous policies.

\subsection{Energy costs for improving smoothness}

In industrial applications, we often look for energy efficiency or smoothness of controllers, besides pure performance. Generally, if two different policies meet the performance criteria, often we would choose the smoother one, because it is more energy efficient and causes less wear and tear on the system.

Therefore, we add a very small but non-zero energy cost to all non-zero actions, which is $0.001$ times the standard step cost. We want to get the policy to move as little as possible, and to select the zero-action as often as possible while still being in the target area \textit{and} still optimizing the pole angle.

The results are listed in table \ref{tab:performance_metrics_improved_policies} as TOESHPQ and a trajectory plot is shown in figure \ref{fig:improving_policy_134_episodes} (c). As can be seen in the plot, the policy learned after 341 episodes visibly moves less than the previous policy, selecting even smaller actions than before. While the $e_{\infty}$ is similar, slightly better, than the previous policy, and the swing-up behavior further improved, selecting the smaller actions comes at a small cost in terms of a slightly higher maximum deviation $e_T$.

A few iterations earlier, after 336 episodes, another policy showed even slightly better balancing performance ($e_{\infty}=0.51$) but a by 5 steps in average worse stabilizing performance.

\subsection{Removing actions from the controller}

\begin{figure}[htbp]
\includegraphics[width=\columnwidth]{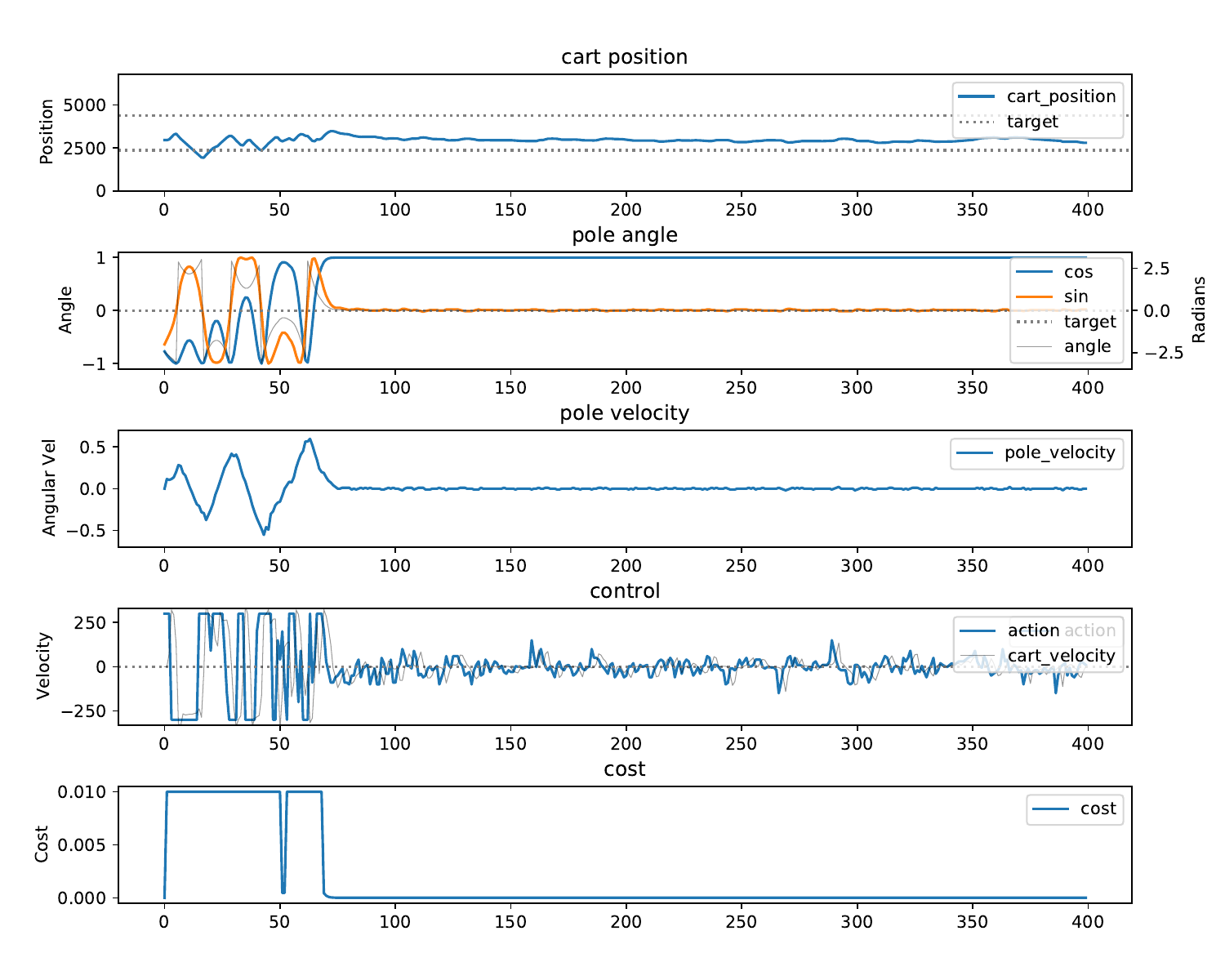}
\caption{Evaluation of policy 341, but here with the two strongest actions (500, -500) removed. The policy is visibly less aggressive in the swing up phase, but still shows the same smooth balancing performance.}
\label{fig:improving_policy_134_episodes_deleted_500}
\end{figure}
As the final experiment, we do the opposite of the earlier experiment where we added actions to an existing policy. We remove the two strongest actions (500, -500) from the controller and evaluate the policy again. Again, update the network and the data in any ways, we just let the actor not consider these two actions when selecting the action with the minimum q-value in the action selection step.

The results are listed in table \ref{tab:performance_metrics_improved_policies} as DEL.341, a trajectory is shown in figure \ref{fig:improving_policy_134_episodes_deleted_500}. Whereas the policy is visibly less aggressive in the swing up phase, it still shows the same smooth balancing performance.

\subsection{Repetition for an uncut video}

We have repeated the learning process from the beginning to record an uncut video of the full learning process.\footnote{The video is available at \url{https://www.youtube.com/watch?v=JghiWOxWa-o}.} For this run, we used shorter episodes of 200 steps each and $epsilon=0$, so that we did not have to run a separate evaluation episode (see discussion in sec. \ref{sec:ablations}). We started with three actions and then, after learning to stabilize and balance, extended the action space to 35 actions, this time adding two additional very strong actions of (-750, 750). At this time we also switched to the time optimal cost function, with shaping in a very narrow target angle margin of $\theta_{\alpha}=0.0005$, a gamma value of $0.999$ and small energy costs for all non-zero actions. The resulting policy swings up even faster than the previous best policy, achieving $N=36$ and looks similar to the previous policies, otherwise.

\section{Discussion}

In revisiting the NFQ algorithm through the lens of modern reinforcement learning, this paper has sought to bridge the gap between foundational RL techniques and contemporary applications in industrial control systems. The introduction of NFQ2.0, a modernized variant of the original NFQ, demonstrates that with integration of current techniques and best practices, NFQ as the vanilla batch Q-learning algorithm of Deep RL can still offer competitive and stable learning performance in low-dimensional real-world control problems, and, thus, is a viable approach for applying deep reinforcement learning in industrial contexts.

Specifically, the use of larger, more modern networks and the switch to continuously updating a single neural network, instead of resetting or replacing it, has shown to significantly enhance the learning process, allowing for faster convergence to good control policies and more reliable policy performance. Furthermore, learning progresses have been found to be more robust and consistent than the original NFQ variant.

Using the batch learning toolset to apply further improvements to the learned policies, such as expanding the action space and dynamically adjusting cost functions, demonstrate the flexibility and adaptability of the growing batch toolset. These enhancements not only improve policy smoothness and performance but also illustrate the potential for iterative and efficient policy refinement in industrial settings. The ability to bootstrap, modify and re-learn policies offline and with minimal additional system interaction is particularly valuable in environments where operational constraints limit extensive experimentation.

In conclusion, NFQ2.0 exemplifies how conceptually simple and computationally efficient approaches can excel when combined with the modern techniques, offering a viable path for applying deep reinforcement learning in industrial settings.

Future work could revisit NFQ-CA, NFQ's classic variant for continuous action spaces, to see if it could also benefit further from the modern techniques.

\bibliographystyle{plainnat}
\bibliography{references}

@inproceedings{riedmiller2005neural,
  title={Neural Fitted Q Iteration - First Experiences with a Data Efficient Neural Reinforcement Learning Method},
  author={Riedmiller, Martin},
  booktitle={Machine Learning: ECML 2005},
  pages={317--328},
  year={2005},
  publisher={Springer},
  series={Lecture Notes in Computer Science},
  volume={3720},
  doi={10.1007/11564096_32}
}

@article{hafner2011reinforcement,
  title={Reinforcement learning in continuous action spaces},
  author={Hafner, Roland and Riedmiller, Martin},
  journal={2011 IEEE Symposium on Adaptive Dynamic Programming and Reinforcement Learning (ADPRL)},
  pages={272--279},
  year={2011},
  publisher={IEEE},
  doi={10.1109/ADPRL.2011.5967376}
}

@inproceedings{lange2010deep,
  title={Deep auto-encoder neural networks in reinforcement learning},
  author={Lange, Sascha and Riedmiller, Martin},
  booktitle={The 2010 International Joint Conference on Neural Networks (IJCNN)},
  pages={1--8},
  year={2010},
  organization={IEEE},
  doi={10.1109/IJCNN.2010.5596468}
}

@article{ormoneit2002kernel,
  title={Kernel-based reinforcement learning},
  author={Ormoneit, Dirk and Sen, {\v{S}}aunak},
  journal={Machine learning},
  volume={49},
  pages={161--178},
  year={2002},
  publisher={Springer},
  doi={10.1023/A:1017928328829}
}

@article{mnih2015human,
  title={Human-level control through deep reinforcement learning},
  author={Mnih, Volodymyr and Kavukcuoglu, Koray and Silver, David and Rusu, Andrei A and Veness, Joel and Bellemare, Marc G and Graves, Alex and Riedmiller, Martin and Fidjeland, Andreas K and Ostrovski, Georg and others},
  journal={Nature},
  volume={518},
  number={7540},
  pages={529--533},
  year={2015},
  publisher={Nature Publishing Group},
  doi={10.1038/nature14236}
}

@inproceedings{lillicrap2015continuous,  
  title={Continuous control with deep reinforcement learning},
  author={Lillicrap, Timothy P and Hunt, Jonathan J and Pritzel, Alexander and Heess, Nicolas and Erez, Tom and Tassa, Yuval and Silver, David and Wierstra Daan},
  booktitle={3rd International Conference on Learning Representations, ICLR 2015},
  year={2015},
  url={https://arxiv.org/abs/1509.02971}
}

@inproceedings{lange2012batch,
  title={Batch reinforcement learning},
  author={Lange, Sascha and Gabel, Thomas and Riedmiller, Martin},
  booktitle={Reinforcement Learning: State-of-the-Art},
  pages={45--73},
  year={2012},
  publisher={Springer},
  series={Adaptation, Learning, and Optimization},
  volume={12},
  doi={10.1007/978-3-642-27645-3_2}
}

@article{gordon1995stable,
  title={Stable fitted reinforcement learning},
  author={Gordon, Geoffrey J},
  journal={Advances in neural information processing systems},
  volume={8},
  year={1995}
}

@article{ernst2005tree,
  title={Tree-Based Batch Mode Reinforcement Learning},
  author={Ernst, Damien and Geurts, Pierre and Wehenkel, Louis},
  journal={Journal of Machine Learning Research},
  volume={6},
  pages={503--556},
  year={2005},
  url={https://www.jmlr.org/papers/v6/ernst05a.html}
}

@inproceedings{lagoudakis2003least,
  title={Least-Squares Policy Iteration},
  author={Lagoudakis, Michail G. and Parr, Ronald},
  booktitle={Advances in Neural Information Processing Systems},
  volume={15},
  pages={1107--1114},
  year={2003},
  publisher={MIT Press},
  url={https://proceedings.neurips.cc/paper/2002/hash/ee8fe9093fbbb687bef15a38facc44d2-Abstract.html}
}

@inproceedings{abdolmaleki2018maximum,
title={Maximum a Posteriori Policy Optimisation},
author={Abbas Abdolmaleki and Jost Tobias Springenberg and Yuval Tassa and Remi Munos and Nicolas Heess and Martin Riedmiller},
booktitle={International Conference on Learning Representations},
year={2018},
url={https://openreview.net/forum?id=S1ANxQW0b},
}

@article{mnih2013playing,
  title={Playing Atari with Deep Reinforcement Learning},
  author={Mnih, Volodymyr and Kavukcuoglu, Koray and Silver, David and Graves, Alex and Antonoglou, Ioannis and Wierstra, Daan and Riedmiller, Martin},
  journal={arXiv preprint arXiv:1312.5602},
  year={2013},
  url={https://arxiv.org/abs/1312.5602}
}

@inproceedings{riedmiller2005pole,
  title={Neural reinforcement learning to swing-up and balance a real pole},
  booktitle={2005 IEEE International Conference on Systems, Man and Cybernetics},
  author={Riedmiller, Martin},
  volume={4},
  pages={3191--3196},
  year={2005},
  organization={IEEE}
}

@article{riedmiller2012tricks,
  title={10 steps and some tricks to set up neural reinforcement controllers},
  author={Riedmiller, Martin},
  journal={Neural Networks: Tricks of the Trade: Second Edition},
  pages={735--757},
  year={2012},
  publisher={Springer}
}

@article{lin1992self,
  title={Self-improving reactive agents based on reinforcement learning, planning and teaching},
  author={Lin, Long-Ji},
  journal={Machine learning},
  volume={8},
  pages={293--321},
  year={1992},
  publisher={Springer}
}

@inproceedings{glorot2010understanding,
  title={Understanding the difficulty of training deep feedforward neural networks},
  author={Glorot, Xavier and Bengio, Yoshua},
  booktitle={Proceedings of the thirteenth international conference on artificial intelligence and statistics},
  pages={249--256},
  year={2010},
  organization={JMLR Workshop and Conference Proceedings}
}

@article{andrychowicz2017hindsight,
  title={Hindsight experience replay},
  author={Andrychowicz, Marcin and Wolski, Filip and Ray, Alex and Schneider, Jonas and Fong, Rachel and Welinder, Peter and McGrew, Bob and Tobin, Josh and Pieter Abbeel, OpenAI and Zaremba, Wojciech},
  journal={Advances in neural information processing systems},
  volume={30},
  year={2017}
}

@article{li2020generalized,
  title={Generalized hindsight for reinforcement learning},
  author={Li, Alexander and Pinto, Lerrel and Abbeel, Pieter},
  journal={Advances in neural information processing systems},
  volume={33},
  pages={7754--7767},
  year={2020}
}

@InProceedings{pmlr-v164-riedmiller22a,
  title = 	 {Collect and Infer -- a fresh look at data-efficient Reinforcement Learning},
  author =       {Riedmiller, Martin and Springenberg, Jost Tobias and Hafner, Roland and Heess, Nicolas},
  booktitle = 	 {Proceedings of the 5th Conference on Robot Learning},
  pages = 	 {1736--1744},
  year = 	 {2022},
  editor = 	 {Faust, Aleksandra and Hsu, David and Neumann, Gerhard},
  volume = 	 {164},
  series = 	 {Proceedings of Machine Learning Research},
  month = 	 {08--11 Nov},
}

\end{document}